%% file: main.tex
\documentclass[final]{cvpr}
\input{setup/package}
\input{setup/macros}
\input{setup/symbols}

\input{setup/graphicspath}

\usepackage{animate}
\usepackage{subfiles} 
\def\cvprPaperID{5075} % *** Enter the CVPR Paper ID here
\def\confYear{CVPR 2021}
\begin{document}

%%%%%%%%% TITLE
\title{Robust Consistent Video Depth Estimation}

\author{
Johannes Kopf\\
Facebook\\
\and
Xuejian Rong\\
Facebook\\
\and
Jia-Bin Huang\\
Virginia Tech
}

% \makeatletter
% \twocolumn[{
% \renewcommand\twocolumn[1][]{#1}
% \maketitle
% \input{figures/teaser}
% }]

\twocolumn[{
\renewcommand\twocolumn[1][]{#1}
\maketitle
\centering \vspace{-10mm}\url{https://robust-cvd.github.io} 
\vspace{10mm}
\input{figures/teaser}
}]

\thispagestyle{empty}
\input{0_abstract}
\input{1_introduction}
\input{2_prior}

\input{3_method}
\input{4_result}

\input{5_conclusion}

{\small
\bibliographystyle{ieee_fullname}
\bibliography{main,cvd-1}
}
% \subfile{sup}
\end{document}

% \onecolumn
% {
% \Large
% \begin{center}
% \textbf{\noindent 3D Photography using Context-aware Layered Depth Inpainting\\Supplementary Material}
% \end{center}
% \maketitle
% }

\title{Robust Consistent Video Depth Estimation \\Supplementary Material}

\author{
      Johannes Kopf\textsuperscript{\dag}
\quad Xuejian Rong\textsuperscript{\dag}
\quad Jia-Bin Huang\textsuperscript{\S}
\vspace{0.5em}\\
\textsuperscript{\dag}Facebook
\qquad \textsuperscript{\S}Virginia Tech
\vspace{0.2em}\\
{\tt\small \textsuperscript{\dag}\{jkopf, xrong\}@fb.com}, \quad
{\tt\small \textsuperscript{\S}jbhuang@vt.edu}
}

% \twocolumn[{
% \renewcommand\twocolumn[1][]{#1}
% \maketitle
% }]
\maketitle

% Main paper
\input{sup_content}

\clearpage
{\small
\bibliographystyle{ieee_fullname}
\bibliography{main,cvd-1}
}

%% file: setup/package.tex
\usepackage{graphicx}
\usepackage{subcaption}
\usepackage{float}
\usepackage[justification=raggedright]{caption}	% makes captions ragged right - thanks to Bryce Lobdell
\usepackage{lscape}                                         % Useful for wide tables or figures.
\usepackage{wrapfig}

\usepackage[lined,ruled,linesnumbered]{algorithm2e}
\usepackage{animate} %% convert frames to video

\usepackage{booktabs}                   % Publication quality tables
\usepackage{multirow}

\usepackage{makecell}

\usepackage{paralist}
\usepackage{enumitem}
\setlist[enumerate]{itemsep=0mm}

\usepackage{bm}                          % Make bold, italic math symbols
\usepackage{epsfig}                      % for figures
\usepackage{graphicx}                  % another package that works for figures
\usepackage{times}
\usepackage{mathptmx}
\usepackage{mathtools}
\usepackage{amssymb,amsmath}   % Short math guide for LaTeX ftp://ftp.ams.org/pub/tex/doc/amsmath/short-math-guide.pdf

\usepackage{units}
\usepackage{color}

\usepackage{comment}

\usepackage{url}  % Hyphenation of URLs.
\usepackage[pagebackref=true,breaklinks=true,letterpaper=true,colorlinks,bookmarks=false]{hyperref}
\usepackage{xspace}
\usepackage[table]{xcolor}
\usepackage{setspace}
\usepackage{grfext}
\PrependGraphicsExtensions*{.jpg,.png,.PNG}

%% file: setup/macros.tex
\usepackage{soul}
\usepackage{svg}

\def\vs{vs.~}                 % against

\def\naive{na{\"i}ve\xspace}
\def\Naive{Na{\"i}ve\xspace}

\DeclareMathOperator*{\argmin}{\arg\!\min} 

\DeclareMathOperator*{\minimize}{minimize}

\DeclareMathAlphabet{\altmathcal}{OMS}{cmsy}{m}{n}

\newlength\paramargin
\newlength\figmargin

\newlength\secmargin
\newlength\figcapmargin
\newlength\tabcapmargin

\newcommand{\changed}{\textcolor{black}} 

\newcommand{\mpage}[2]
{
\begin{minipage}{#1\linewidth}\centering
#2
\end{minipage}
}

\newcommand{\topic}[1]
{
\vspace{1mm}\noindent\textbf{#1}
}

\newcommand{\figref}[1]{Figure~\ref{fig:#1}} 
\newcommand{\tabref}[1]{Table~\ref{tab:#1}}

\long\def\ignorethis#1{}
\newcommand {\jiabin}[1]{{\color{cyan}\textbf{Jia-Bin: }#1}\normalfont}
\newcommand {\johannes}[1]{{\color{red}\textbf{Johannes: }#1}\normalfont}

\newcommand{\unsure}[1]{{\sethlcolor{yellow}\hl{#1}}}

\newcommand{\tb}[1]{\textbf{#1}}

\newbox\jsavebox%
\newcommand{\jsubfig}[2]{%
	\sbox\jsavebox{#1}%
	\parbox[t]{\wd\jsavebox}{\centering\usebox\jsavebox\\#2}%
	}

\makeatletter
\newcommand{\providelength}[1]{%
  \@ifundefined{\expandafter\@gobble\string#1}
   {% if #1 is undefined, do \newlength
    \typeout{\string\providelength: making new length \string#1}%
    \newlength{#1}%
   }
   {% else check whether #1 is actually a length
    \sdaau@checkforlength{#1}%
   }%
}

\newcommand{\sdaau@checkforlength}[1]{%
  \edef\sdaau@temp{\expandafter\sdaau@getfive\meaning#1TTTTT$}%
  \ifx\sdaau@temp\sdaau@skipstring
    \typeout{\string\providelength: \string#1 already a length}%
  \else
    \@latex@error
      {\string#1 illegal in \string\providelength}
      {\string#1 is defined, but not with \string\newlength}%
  \fi
}
\def\sdaau@getfive#1#2#3#4#5#6${#1#2#3#4#5}
\edef\sdaau@skipstring{\string\skip}
\makeatother

%% file: setup/symbols.tex
\def\xi{\mathbf{x}_i}

%% file: setup/graphicspath.tex
\graphicspath{{figure}, {example}}

%% file: figures/teaser.tex
\providelength\height%
\setlength\height{8.5cm}%
\providelength\width%
\setlength\width{8cm}%
\providelength\liftup%
\setlength\liftup{-1mm}%

\vspace{-0.8cm}

\ignorethis{
\begin{center}
\jsubfig{\includegraphics[width=\width]{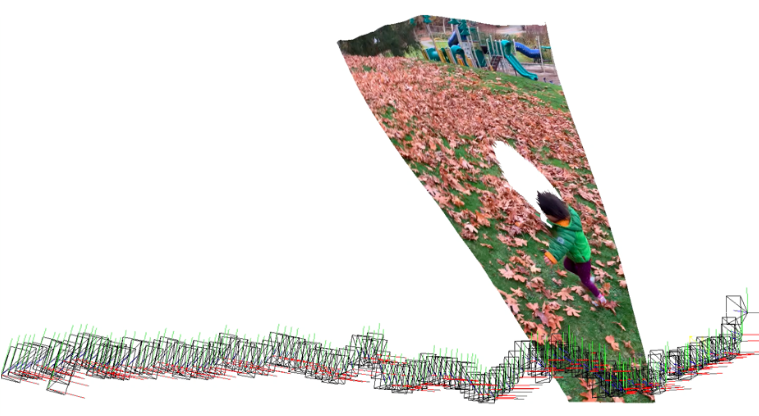}}{}
\hfill%
\jsubfig{\includegraphics[width=\width]{figures/teaser/teaser_1.png}}{}\\
\jsubfig{\includegraphics[width=0.66\width]{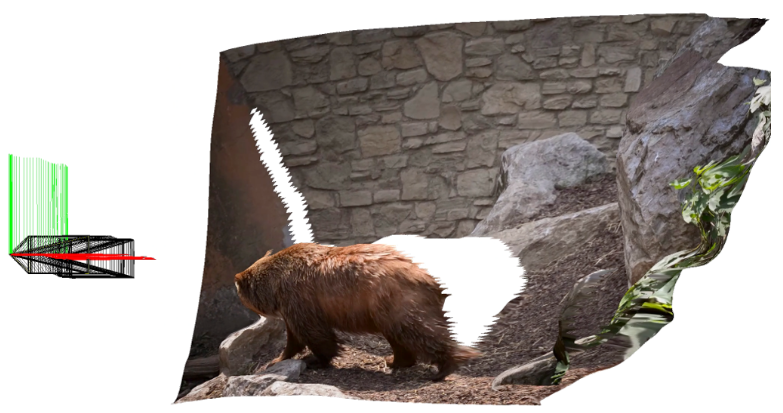}}{}
\hfill%
\jsubfig{\includegraphics[width=0.66\width]{figures/teaser/teaser_2.png}}{}
\hfill
\jsubfig{\includegraphics[width=0.66\width]{figures/teaser/teaser_2.png}}{}
\captionof{figure}{
\tb{Robust consistent video depth estimation.} Our method estimates a smooth camera trajectory and detailed and stable dense depth map on challenging hand-held cellphone videos.
\jiabin{Two long sequences on the two. Three short ones below.}
}
\label{fig:strip}
\end{center}
}

\includegraphics[width=\textwidth]{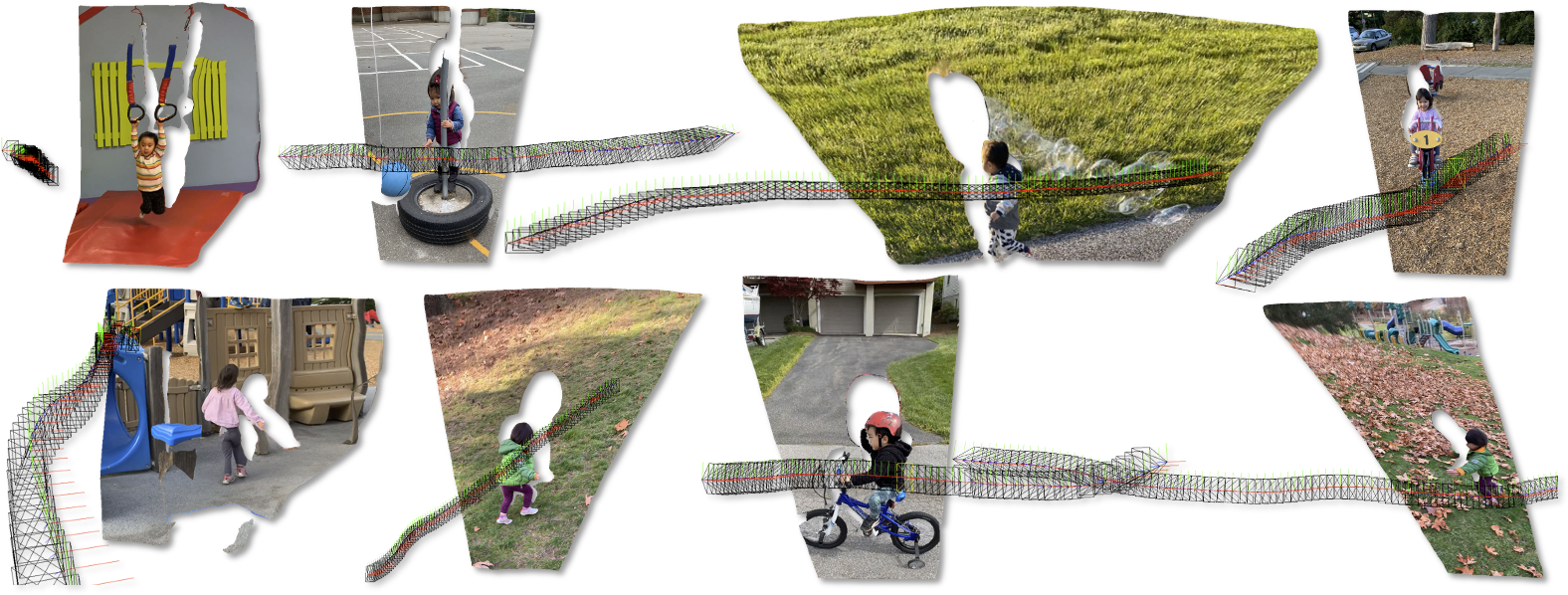}
\captionof{figure}{
\tb{Robust consistent video depth estimation of dynamic scenes.} Our method estimates a smooth camera trajectory and detailed and stable dense depth map on challenging hand-held cellphone videos. Our method supports both still (static) and dynamic camera motion.
}
\vspace{2mm}

%% file: 0_abstract.tex
\begin{abstract}
We present an algorithm for estimating consistent dense depth maps and camera poses from a monocular video.
We integrate a learning-based depth prior, in the form of a convolutional neural network trained for single-image depth estimation, with geometric optimization, to estimate a smooth camera trajectory as well as detailed and stable depth reconstruction.
Our algorithm combines two complementary techniques: (1) flexible deformation-splines for low-frequency large-scale alignment and (2) geometry-aware depth filtering for high-frequency alignment of fine depth details.
In contrast to prior approaches, our method does not require camera poses as input and achieves robust reconstruction for challenging hand-held cell phone captures containing a significant amount of noise, shake, motion blur, and rolling shutter deformations.
Our method quantitatively outperforms state-of-the-arts on the Sintel benchmark for both depth and pose estimations and attains favorable qualitative results across diverse wild datasets.

\end{abstract}

%% file: 1_introduction.tex
\section{Introduction}
\label{sec:intro}

Dense per-frame depth is an important intermediate representation that is useful for many video-based applications, such as 3D video stabilization \cite{Liu2009}, augmented reality (AR) and special video effects \cite{valentin2018depth,Luo-VideoDepth-2020}, and for converting videos for virtual reality (VR) viewing \cite{Huang2017}.
However, estimating accurate and consistent depth maps for casually captured videos still remains very challenging.
Cell phones contain small image sensors that may produce noisy images, especially in low lighting situations.
They use a rolling shutter that may result in wobbly image deformations.
Hand-held captured casual videos often contain camera shake and motion blur, and dynamic objects, such as people, animals, and vehicles.
In addition to all these degradations, there exist well-known problems for 3D reconstruction that are not specific to video, including poorly textured image regions, repetitive patterns, and occlusions.

\emph{Traditional} algorithms for dense reconstruction that combine Structure from Motion (SFM) and Multi-view Stereo (MVS) have difficulties dealing with these challenges.
The SFM step suffers from the limitations of accuracy and availability of correspondence and often fails entirely, as explained below, preventing further processing.
Even when SFM succeeds, the MVS reconstructions typically contain a significant amount of holes and noises.

\emph{Learning-based} algorithms \cite{li2018megadepth,li2019learning,lasinger2019towards} are better equipped to handle with this situation.
Instead of matching points across frames and geometric triangulation they employ priors learned from diverse training datasets.
This enables them to handle many of the challenging situations aforementioned.
However, the estimated depth is only defined \emph{up to scale}, and, while plausible, is not necessarily accurate, i.e., it lacks geometric consistency.

\emph{Hybrid} algorithms \cite{Luo-VideoDepth-2020,liu2019neural,yoon2020novel,teed2020deepv2d} achieve desirable characteristics of both approaches by combining learnt priors with geometric reasoning. 
These methods often assume precise per-frame camera poses as auxiliary inputs, which are typically estimated with SFM.
However, SFM algorithms are not robust to the issues described above.
In such situations, SFM might fail to register all frames or produce outlier poses with large errors.
As a consequence, hybrid algorithms work well when the pose estimation succeeds and fail catastrophically when it does not.
This problem of \emph{robustness} makes these algorithms unsuitable for many real-world applications, as they might fail in unpredictable ways.
Recently, a hybrid approach proposed in the DeepV2D work \cite{teed2020deepv2d} attempts to interleave pose and depth estimations in inference for an ideal convergence, which performs reasonably well on static scenes but still does not prove the capability of tackling dynamic scenes.

%\johannes{Do we need to say anything about DeepV2D here?} 
%\jiabin{Yep, DeepV2D is also kind of hybrid because they use FRCN to initialize their depth estimation.}

We present a new algorithm that is more robust and does \emph{not} require poses as input.
Similar to Luo et al.~\cite{Luo-VideoDepth-2020}, we leverage a convolutional neural network trained for single-image depth estimation as a depth prior and optimize the \emph{alignment} of the depth maps.
However, their test-time fine-tuning formulation requires a pre-established geometric relationship between matched pixels across frames, which, in turn, requires precisely calibrated camera poses and per-frame depth scale factors.
In contrast, we \emph{jointly optimize} extrinsic and intrinsic camera pose parameters as well as \emph{3D alignment} of the estimated depth maps using continuous optimization.
\Naive alignment using rigid-scale transformations does not result in accurate poses because the \emph{independently} estimated per-frame depth maps usually contain random inaccuracies. These further lead to misalignment, which inevitably imposes noisy errors onto the estimated camera trajectory.
We resolve it by turning to a more flexible deformation model, using spatially-varying splines.
They provide a more exact alignment, which, in succession, results in smoother and more accurate trajectories.

The spline-based deformation achieves accurate \emph{low-frequency} alignment.
To further improve \emph{high-frequency} details and remove residual jitter, we use a geometry-aware depth filter.
This filter is capable of bringing out fine depth details, rather than blurring them because of the precise alignment from the previous stage.

As shown in previous work, the learning-based prior is resilient to moderate amounts of dynamic motion.
We make our method even more robust to large dynamic motion by incorporating automatic segmentation-based masks to relax the geometric alignment requirements in regions containing people, vehicles, and animals.

We evaluate our method qualitatively (visually) by processing \emph{all} 90 sequences from the DAVIS dataset (originally designed for dynamic video object segmentation) \cite{pont20172017} and comparing against previous methods (of which many fail).
We further, evaluate quantitatively on the 23 sequences from the Sintel dataset \cite{ButlerSintel2012}, for which ground truth pose and depth are available.

% The source code of our method is available at \url{http://github.com/.../} (after the submission phase)

%\johannes{We should probably cite Instant 3D Photography \cite{hedman2018instant} somewhere, since our formulation resembles it.}

%\xuejian{We may need a short paragraph here to quickly summarize the main contributions?}

%% file: 2_prior.tex
\section{Related Work}
\label{sec:related}

\topic{Multi-view stereo.}
Multi-view stereo (MVS) algorithms estimate depth from a collection of images captured from different viewpoints \cite{seitz2006comparison,furukawa2015multi}.
Geometry-based MVS systems (e.g., COLMAP ~\cite{schonberger2016structure}) follow the incremental Structure-from-Motion (SFM) pipeline (correspondence estimation, pose estimation, triangulation, and bundle adjustment).
Several learning-based methods further improve the reconstruction quality by fusing classic MVS techniques (e.g., cost aggregation and plane-sweep volume) and data-driven priors \cite{ummenhofer2017demon,huang2018deepmvs,yao2018mvsnet,im2019dpsnet,kusupati2020normal}.
In contrast to MVS algorithms that assume a \emph{static} scene, our work aims to reconstruct fully dense depth from a dynamic scene video.

\topic{Single-image depth estimation.}
In recent years we have witnessed rapid progresses on \emph{supervised} learning-based single-image depth estimation \cite{eigen2014depth,eigen2015predicting,laina2016deeper,liu2015learning,fu2018deep}.
As diverse training images with the corresponding ground truth depth maps are difficult to obtain, existing work explores training models using 
synthetic datasets \cite{mayer2016large}, crowd-sourced human annotations of relative depth~\cite{chen2016single} or 3D surfaces \cite{chen2020oasis}, pseudo ground truth depth maps from internet images/videos \cite{li2018megadepth,li2019learning,chen2019learning}, or 3D movies \cite{lasinger2019towards,wang2019web}. 
Another research line focuses on \emph{self-supervised} approaches for learning single-image depth estimation models.
Specific examples include learning from stereo pairs \cite{godard2017unsupervised,guo2018learning,gordon2019depth,watson2019self} or monocular videos~\cite{zhou2017unsupervised,vijayanarasimhan2017sfm,dai2019self,zou2018df,yin2018geonet,qi2018geonet,ranjan2019competitive}.
Most self-supervised learning methods minimize photometric reprojection errors (computed from the estimated depth and pose) and do not account for dynamic objects in videos.
Several methods alleviate this problem by masking out dynamic objects~\cite{zou2018df}, modeling the motion of individual objects \cite{casser2019depth} or estimating dense 3D translation field~\cite{li2020unsupervised}.
We use the state-of-the-art single-image depth estimation method~\cite{lasinger2019towards} to obtain an initial dense depth map for each video frame. 
While these depth maps are plausible when viewed individually, they are \emph{not} geometrically consistent across different frames. 
Our work aims to produce geometrically consistent camera poses and dense depth for a video.

\topic{Video-based depth estimation.}
Several methods integrate camera motion estimation and multi-view reconstruction from a pair of frames \cite{ummenhofer2017demon} or multiple frames \cite{zhou2018deeptam,bloesch2018codeslam,valentin2018depth}.
However, these methods work well only on a static scene. 
To account for moving objects, a line of work use motion segmentation~\cite{karsch2014depth,ranftl2016dense} or semantic instance segmentation~\cite{casser2019depth} to help constrain the depth of moving objects. 
State-of-the-art learning-based video depth estimation approaches can be grouped into two tracks: (1) MVS-based methods and (2) hybrid methods.
MVS-based methods improve the conventional SFM and MVS workflow using differentiable pose/depth modules~\cite{teed2020deepv2d} or explicit modeling depth estimation uncertainty~\cite{liu2019neural}.
Both methods \cite{liu2019neural,teed2020deepv2d} estimate depth based on the cost volume constructed by warping nearby frames to a reference viewpoint.
These methods may produce erroneous depth estimation and fail to generate accurate camera trajectories for dynamic scenes.
Hybrid methods integrate single-view depth estimation models and multi-view stereo for achieving \emph{geometrically consistent} video depth, either through fusing depth predictions from single-view and multi-view~\cite{yoon2020novel} or fine-tuning single-view depth estimation model to satisfy geometric consistency~\cite{Luo-VideoDepth-2020}. 
While impressive results have been shown, the methods \cite{yoon2020novel,Luo-VideoDepth-2020} assume that precise camera poses are available as input and thus are not applicable for challenging sequences where existing SFM/MVS systems fail. 
% Relationship
Our method also leverages a pretrained single-view depth estimation model.
Unlike \cite{yoon2020novel,Luo-VideoDepth-2020}, we {jointly} optimize camera poses and 3D deformable alignment of the depth maps and thus can handle a broader range of challenging videos of highly dynamic scenes.

\topic{Visual odometry.}
% Related work
Visual Odometry (VO) or Simultaneously Localization And Mapping (SLAM) aim to estimate the relative camera poses from image sequences \cite{nister2004visual,scaramuzza2011visual}.
Conventional geometry-based methods~\cite{engel2014lsd,engel2017direct,mur2017orb,ranftl2016dense,newcombe2011dtam} can be grouped into the four categories depending on using \emph{direct} (feature-less) \vs \emph{indirect} (feature-based) methods and \emph{dense} or \emph{sparse} reconstruction.
While significant progress has been made, applying VO and SLAM for generic scenes remains challenging~\cite{yang2018challenges}.
Recent years, numerous learning-based approaches have been proposed to tackle these challenges either via supervised~\cite{ummenhofer2017demon,wang2017deepvo,zhou2018deeptam,gradslam} or self-supervised learning~\cite{zhou2017unsupervised,li2018undeepvo,zhan2018unsupervised,godard2019digging,sheng2019unsupervised,xue2019beyond,zou2020learning,yang2020d3vo}.
% Relationship
Similar to the existing VO/SLAM methods, our work also estimates both camera poses from a monocular video. 
Unlike prior work, our primary focus lies in estimating geometrically consistent dense depth reconstruction for \emph{dynamic scenes}. 

% \johannes{\unsure{Add section about TEMPORAL CONSISTENCY.}}

\topic{Temporal consistency.} 
Per-frame processing often leads to temporal flickering results. 
Enforcing temporal consistency of a output video has been explored in many different applications, including style transfer~\cite{chen2017coherent}, video completion~\cite{huang2016temporally}, video synthesis~\cite{wang2018video}, or as post-processing step~\cite{lang2012practical,bonneel2015blind,lai2018learning}.
For video depth estimation, temporal consistency can either be explicitly constrained by optical flow~\cite{karsch2014depth} or implicitly applied using recurrent neural networks~\cite{Patil2020dont}.
% Relationship
Our 3D depth filter is similar to that of \cite{lang2012practical} because we also filter the depth maps \emph{across} time along the flow trajectory. 
Our approach differs in that our method is \emph{geometry-aware}.

%% file: 3_method.tex
\def\ij{{i \rightarrow j}}
\def\ji{{j \rightarrow i}}

\def\fij{f_\ij}
\def\p{(\hspace{-0.2mm}p\hspace{-0.2mm})}
\def\q{(\hspace{-0.2mm}q\hspace{-0.2mm})}

\newcommand{\x}[1]{x_\textrm{\tiny #1}}

\def\depth{d}
\def\chainflow{\tilde{f}}

\def\finv{u_i^{-1}}

\def\alParams{\theta^{\!\textit{cam}}}
\def\ftParams{\theta^{\!\textit{depth}}}

\def\Mflow{m^\textit{flow}}
\def\Mdyn{m^\textit{dyn}}

\def\Espatial{\altmathcal{E}_\ij^\textit{spatial}}
\def\Edepth{\altmathcal{E}_\ij^\textit{depth}}
\def\Ereg{\altmathcal{E}_i^\textit{regularization}}
\def\Edeform{\altmathcal{E}_i^\textit{deform}}
\def\Efocal{\altmathcal{E}_i^\textit{focal}}

\def\Lrepro{\altmathcal{L}^\textit{repro}}
\def\Ld{\altmathcal{L}^\textit{sim}}
\def\Ldeform{\altmathcal{L}^\textit{deform}}
\def\Lspatial{\altmathcal{L}^\textit{spatial}}
\def\Leuclidean{\altmathcal{L}^\textit{euclidean}}
\def\Ldepth{\altmathcal{L}^\textit{depth}}
\def\Lratio{\altmathcal{L}^\textit{ratio}}
\def\Ldisparity{\altmathcal{L}^\textit{disparity}}
\def\Lfocal{\altmathcal{L}^\textit{focal}}

\section{Overview}
\label{sec:overview}

\input{figures/overview.tex}

Our approach builds on the formulation established in the Consistent Video Depth Estimation (CVD) paper by Luo et al.~\cite{Luo-VideoDepth-2020}, so let us start by recapping it, first.
They iteratively fine-tune the weights of a CNN trained for single-image depth estimation until it learns to satisfy the geometry of a particular scene in question.
To assess the progress against this goal, they relate pairs of images geometrically using \emph{known} camera parameters (extrinsic and intrinsic, as well as per-frame depth scale factors).
More precisely, their algorithm compares the \emph{3D reprojection} of pixels from one image to the other with the corresponding \emph{image-space motion}, computed by an optical flow method.
The reprojection error is back-propagated to the network weights, so that it reduces over the course of the fine-tuning (see details in Section~\ref{sec:depth_estimation}).
This results in a very detailed and temporally consistent, i.e., flicker-free, depth video.

However, one key limitation of their approach is that \emph{precise} camera parameters are needed as input, which are computed with SFM in their case.
Unfortunately, SFM for video is a challenging problem in itself, and it frequently fails; for example, when a video does not contain sufficient camera motion, or in the presence of dynamic object motion, or in numerous other situations, as we explained earlier.
In such cases, it might either fail entirely to produce an output, or fail to register a subset of the frames, or it might produce erroneous (outlier) camera poses.
Inaccurate poses have a strong degrading effect on the CVD optimization, as shown in their paper \cite{Luo-VideoDepth-2020}.
The key contribution of our paper is to remove this limitation, by replacing the test-time fine-tuning with \emph{joint optimization} of the camera parameters and depth alignment.

We show in Section~\ref{sec:pose_estimation} that the same formulation can optimize the camera poses as in CVD.
%While the network weights that control the CNN depth estimates are best optimized using standard training algorithms (i.e., stochastic gradient descent), these methods are less suitable for the camera parameters, since they are more tightly coupled:
%changing the pose of one camera directly and strongly affects all other cameras, since they are chained together.
%These parameters care more suitably optimized using global continuous optimization methods.
However, one complication is that the pose optimization only works well when we have precise depth, similarly to how depth fine-tuning only works when the poses are accurate (Section~\ref{sec:joint_opt_finetune}).
If the depth is not accurate, which is the case in our setting, \emph{misalignments} in the depth impose themselves as noisy errors onto the resulting camera pose trajectory.

We resolve this problem by improving the ability of the camera optimization to align the depth estimates, despite their initial inaccuracy (Section~\ref{sec:joint_opt_flex}).
Specifically, we achieve this by replacing the per-frame camera scale with a more flexible spatially-varying transformation, i.e., a bilinear spline.
%A similar approach has been recently used for panoramic stitching~\cite{hedman2018instant}, although not in a joint depth and pose optimization loop.
The improved alignment of the depth estimates enables computing smoother and more accurate pose trajectories.

The joint pose estimation and deformation resolves \emph{low-frequency} inconsistencies in the depth maps.
We further improve \emph{high-frequency} alignment using a geometry-aware depth filter (Section~\ref{sec:filter}).
This filter low-pass filters the \emph{reprojected} depth along flow trajectories.
Because the input to the filter is well-aligned, due to the deformation, the filter resolves fine details, rather than blurring them.

\section{Method}
\label{sec:method}

\input{figures/opt.tex}

\subsection{Pre-processing}

We share some of the preprocessing steps with CVD~\cite{Luo-VideoDepth-2020}.
However, importantly, we do \emph{not} need to compute COLMAP~\cite{schonberger2016structure}, which considerably improves the robustness of our method.

\setlength{\columnsep}{3mm}%
\begin{wrapfigure}[3]{r}[0cm]{0.5\columnwidth}%
\vspace{-1mm}\includegraphics[width=0.5\columnwidth]{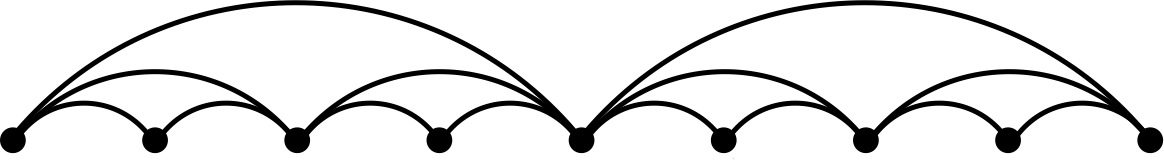}%
\end{wrapfigure}%
In order to lower the overall amount of computation in the pairwise optimization below, we subsample a set of frame pairs spanning temporally near and distant frames:
\begin{equation}
P = \Big\{
(i,~j)~\!\Big\vert
\left| i - j \right| = k,~i~\textrm{mod}~k = 0,~k = 1, 2, 4, \ldots
\Big\}.
\end{equation}

For each frame pair $(i, j) \in P$ we compute a dense optical flow field $\fij$ (mapping a pixel in frame $i$ to its corresponding location in frame $j$) using RAFT \cite{Teed2020}), as well as a binary mask $\Mflow_\ij$ that indicates forward-backward consistent flow pixels.
Please refer to \cite{Luo-VideoDepth-2020} for details about these preprocessing steps.
We also compute a binary segmentation mask $\Mdyn_i$ using Mask R-CNN that indicates likely-dynamic pixels (belonging to the categories ``person'', ``animal'', or ``vehicle'').
%\jiabin{In previous paragraphs we mentioned AUTOMATIC mask generation. Seems in conflict with this?}

\subsection{Depth Estimation}
\label{sec:depth_estimation}

In this section, we establish CVD~\cite{Luo-VideoDepth-2020} from a technical point of view and some notation, and in the subsequent sections, we will then present our method.

Let $p$ be a 2D pixel coordinate.
We can lift it into a 3D coordinate $c_i\p$ in frame $i$'s \emph{3D camera coordinate} system:
\begin{equation}
c_i\p = s_i ~ \depth_i\p \, \tilde{p}.
\label{eq:ci}
\end{equation}
Here, $s_i$ is the per-frame scale coefficient, and $\depth_i$ is the CNN-estimated depth map, and $\tilde{p}$ is the homogeneous-augmented pixel coordinate, i.e., $[p_x,\,p_y,\,1]^\top$.

We can also project this 3D point into the camera coordinate system of \emph{another} frame $j$:
\begin{equation}
c_\ij\p =
K_j \, R_j^\top \Big(
R_i \, K_i^{-1} c_i\p + t_i - t_j
\Big)
\label{eq:cij}
\end{equation}
Here, $R_i, R_j$ and $t_i, t_j$ and $K_i, K_j$ are the rotation, translation, and intrinsics of frames $i$ and $j$, respectively.

The objective that CVD optimizes is a \emph{reprojection loss} for every pixel (with valid flow) in every frame pair:
\begin{equation}
\argmin_{\ftParams}
~
\sum_{(i, j) \in P\vphantom{\Mflow_\ij}}
~
\sum_{p \in \Mflow_\ij}
~
\Lrepro_\ij\!\p,~~~~~\textrm{s.t. fixed }\alParams
\label{eq:obj}
\end{equation}
The optimization variables, $\ftParams$, are the network weights of the depth CNN, and the fixed camera parameters are  $\alParams = \left\{ R_i, t_i, K_i, s_i \right\}$.

The reprojection loss is defined as follows:
\begin{equation}
\Lrepro_\ij\!\p =
\Ld_j \!
\Big( \hspace{-2.7mm}
\underbrace{c_\ij \p}_\textrm{\scriptsize 3D-reprojection}\hspace{-2mm} ,~\,
c_j \! \big(\hspace{-3.5mm}\underbrace{f_\ij\p}_\textrm{\scriptsize Flow-reprojection} \hspace{-3.5mm}\big)
\!\Big),
\label{eq:repro}
\end{equation}
i.e., it reprojects the pixel into the other frame's 3D camera coordinate system using (1) 3D geometry and (2) optical flow and measures the similarity of the two resulting 3D points.
The exact form of the reprojection similarity loss $\Ld$ is not critical for the overall understanding of the algorithm, so we will defer its definition to below.

\subsection{Pose Optimization}
\label{sec:pose_estimation}

As mentioned before, the camera parameters $\alParams$ are needed for the geometric reprojection mechanics, and it is critically important that they are precise.
Otherwise, the depth optimization converges to poor results (Figure~\ref{fig:opt}d).
It would be desirable to have a more reliable way to obtain poses for our application than with SFM.

When examining Eq.~\ref{eq:obj} we notice that we can actually use this equation to compute the camera parameters if we reverse the role of $\ftParams$ and $\alParams$, i.e., \emph{fixing} $\ftParams$ (assuming that we know them) and \emph{optimizing} $\alParams$.
This modified equation resembles the triangulation in bundle adjustment, but it can be more robustly solved since the depth of matched image points does not need to be estimated since it is known (up to scale).
However, this time it is the \emph{depth} that needs to be known precisely for this to work well.
In the depth maps do not agree with each other, these misalignment errors will manifest as noisy errors in the estimated camera parameters (Figure~\ref{fig:opt}c).

\subsection{Pose and Depth Optimization with Fine-tuning}
\label{sec:joint_opt_finetune}

What about optimizing \emph{both} quantities, $\alParams$ and $\ftParams$, jointly?
One problem is that these two quantities are best optimized with different kinds of machinery.
$\ftParams$ is best optimized using standard training algorithms for CNNs, i.e., SGD.
For $\alParams$, however, that is not a good fit, since changes to one parameters have far-reaching influence, as the poses are chained in a trajectory.
Global continuous optimization is a better solution for $\alParams$ and converges faster and more stably.
We can optimize both quantities by alternating between optimizing depth and pose, each with their respective best optimization algorithm while keeping the other quantity fixed.

However, another significant problem is the sensitivity to the accuracy of the particular fixed parameters, as explained before.
Starting with the initially inaccurate depth estimate will result in noisy poses (Figure~\ref{fig:opt}c), because the misalignment errors will ``push'' the camera pose variables in erratic ways.
The jittery poess will, in the next step, degrade the depth estimate further.
The algorithm does not converge to a good solution (Figure~\ref{fig:opt}d).

\vspace{-0.1em}

\subsection{Pose and Depth with Flexible Deformation}
\label{sec:joint_opt_flex}

Our solution to this apparent dilemma is to improve the \emph{depth alignment} in the pose estimation.
We achieve this by injecting a smooth and flexible spatially-varying \emph{depth deformation model} into the alignment.
More precisely, we replace the depth scale coefficients $s_i$ in Eq.~\ref{eq:obj} with a spatially-varying bilinear spline:
\begin{equation}
\varphi_i\p = \sum_k b_k\!\p \, s_i^k.
\end{equation}
Here, the $s_i^k$ are scale factors that are defined on a regular grid of ``handles'' across the image.
$b_k\!\p$ are bilinear coefficients, such that within a grid cell the four surrounding handles of a pixel $p$ are bilinearly interpolated.

After this change the depth maps align better and will not impose jittery errors onto the estimated camera trajectory anymore (Figure~\ref{fig:opt}e).
In addition, this algorithm is considerably faster since we do not need to iterate between pose optimization and fine-tuning.

\subsection{Geometry-aware Depth Filtering}
\label{sec:filter}

\input{figures/filter.tex}

The flexible deformation $\varphi_i$ achieves a low-frequency alignment of the depth map, i.e., removing any large-scale misalignments.
But what about fine depth details?
We tried using alternating pose-depth optimization, as described in Section~\ref{sec:joint_opt_finetune}, with flexible deformation.
This works and improves both the depth and poses slightly, but it does not reach the same level of quality that CVD achieves when using precise SFM pose as input.
Instead, it converges quickly to a configuration where both depth and pose alignment are well-satisfied, but depth details smooth out considerably.

We solve this problem instead with a spatio-temporal depth filter that follows flow trajectories.
Importantly, the filter is geometry-aware in the sense that it transforms the depths from other frames using the reprojection mechanics in Equations~\ref{eq:ci}-\ref{eq:cij}:
\begin{equation}
\depth_i^\textit{final} \! \p =
\sum_{q \in N\p}
\sum_{j=i-\tau}^{i+\tau}
z_\ji \! \Big( \!\chainflow_\ij \q \!\Big)\,
w_\ij \q.
\end{equation}
Here, \changed{$N\p$ refers to a $3 \times 3$ neighborhood centered around the pixel $p$, $\tau=4$,} $z_\ji$ is the scalar z-component of $c_\ji$ (i.e., the reprojected depth), and $\tilde{f}$ is the flow between far frames obtained by \emph{chaining} flow maps between consecutive frames. The weights $w_\ij\q$ make the filter edge-preserving by reducing the influence os samples across depth edges:
\begin{equation}
w_\ij \q =
\exp \!\!\! \Bigg( \!\!\!\!
-\lambda_f
\Lratio \!\bigg( \!\!
c_i \p,~
c_\ji \Big( \chainflow_\ij \q \Big) \!\!\!
\bigg) \!\!\!\!
\Bigg).
\label{eq:filter_weights}
\end{equation}
$\lambda_f=3$ adjusts the strength of the filter, and $\Lratio$ measures the similarity of the reference and the reprojected pixel in camera coordinates (the definition is given in the implementation details below).

The improved provided by the depth filter can be seen in Figure~\ref{fig:opt}f, and more results in Figure~\ref{fig:filter}.

\vspace{-0.1em}

\subsection{Implementation Details}

\topic{\changed{Precomputation.}}
\changed{
For MiDaS we downscale the image to $384$ pixels on the long side (the default resolution in their code).
For RAFT we downscale the image to $1024$ pixels on the long side, to lower memory requirements.
In both cases we adjust the short image side according to image aspect ratio, while rounding to the nearest multiple of 16 pixels which is the alignment requirement of the respective CNNs.
We use the pretrained \texttt{raft-things.pth} from the RAFT project page \cite{Teed2020}, which was trained on FlyingChairs and FlyingThings.
}

\topic{Pose optimization.}
We use the \changed{\texttt{SPAR\-SE\_NOR\-MAL\_CHO\-LESKY} solver from the Ceres library} \cite{Ceres} to solve the camera pose estimation problem.
In order to reduce the computational complexity, we do not include every pixel in the optimization but instead, subsample a set of pairwise matches from the flow fields (so that there is a minimum distance of at least 10 pixels between any pair of matches).
%using a Poisson disk pattern, so that there is a minimum distance of at least 10 pixels between any pair of matches). \jiabin{cite Poisson disk pattern?}
Since Eq.~\ref{eq:obj} assumes a static scene, we exclude any points matches within the $\Mdyn$ mask.

Since the objective is non-convex optimization, it is somewhat sensitive to local minima in the objective.
%\jiabin{non-linear can still have global minima. I think here is about non-convexity?}
%
We alleviate that problem by first optimizing a $1\times1$ grid (similar to the original $s_i$ scalar coefficients), and then subdividing it in four steps until a grid resolution of $17\times10$ is reached (always using 17 for the long image dimension, and adjusting the short image dimension according to the aspect ratio).
After every step, we optimize until convergence and use the result as initialization for the next step.

Since the scale of the depth maps estimated by the CNN is arbitrary, we initialize the scale of the first frame so that the median depth is 1, and use the same scale for all other frames.

%For a video of \unsure{XXX} frames, all pose optimization steps together take \unsure{XX} seconds on a \unsure{XXX} CPU.

\topic{Pose regularization.}
To encourage smoothness in the deformation field, we add a loss that penalizes large differences in neighboring grid values:
\begin{equation}
\Ldeform =
\sum_i
\sum_{(k, r) \in N}
\Big\|
s_i^k - s_i^r
\Big\|_2^2
\max(\!w_i^k\!,w_i^r),
\end{equation}
\changed{where N refers to the set of all vertically and horizontally neighboring vertices.}
The weights are set to encourage more smoothness in parts of the image that are masked by $\Mdyn$, since there are no point matches in these regions and they are unconstrained otherwise.
\begin{equation}
w_i^k = \lambda_1 + \lambda_2
\sum_p \Mdyn_i\!\p \, b_k\p
\end{equation}
$\lambda_1 = 0.1,~\lambda_2 = 10$ are balancing coefficients.
The $\lambda_2$ term computes the fraction of dynamic pixels in the handle's influence region.

We use the following form for the intrinsic matrices:
\begin{equation}
K_i =
\left[\begin{array}{@{\mkern0mu}l@{\mkern0mu}}
u_i\\%
\hspace{4mm}u_i\\%
\hspace{8.9mm}1%
\end{array}\right]\!\!,
\end{equation}
% \begin{equation}
%     \text{diag}(u_i, u_i, 1),
% \end{equation}
i.e., the only degree of freedom is the focal length $u_i$.

We, further, add a small bias,
\begin{equation}
\Lfocal = \sum_i \left( u_i - \hat{u} \right)^2\!,
\end{equation}
where $\hat{u} = 0.35$ (corresponds to $\sim\!\!40^\circ$ field of view).

\ignorethis{
\topic{Depth optimization.}
We use the Midas network \cite{lasinger2019towards} to compute $d_i$, and initialize $\ftParams$ with the author's pre-trained weights.
We optimize the weights using standard back-propagation with the ADAM optimizer \cite{kingma2014adam}.
We use a batch size of \unsure{XX} and a learning rate of 1e-6.
An epoch is defined by one pass over all frame pairs in $P$.
For a video of \unsure{XXX} frames, a single epoch of training on \unsure{X} NVIDIA \unsure{XXX} GPUs takes \unsure{XX} minutes.
}

\topic{Reprojection loss.}
In Section~\ref{sec:depth_estimation} we omitted the definition of the reprojection similarity loss $\Ld$.
A \naive way would be to simple measure the Euclidean distance
\begin{equation}
\Leuclidean\!(a, b) = \big\| a - b \big\|_2^2.
\end{equation}
However, this biases the solution toward small depths.
Shrinking the whole scene to a point would achieve a minimum.

To prevent this, Luo et al.~\cite{Luo-VideoDepth-2020} use a split loss where they measure the spatial component $\Lspatial$ in image space
\begin{equation}
\Lspatial\!(a, b) = \Big\| \tfrac{a_{xy}}{a_z} - \tfrac{b_{xy}}{b_z}  \Big\|_2^2\,,
\end{equation}
and the depth $\Ldepth$ component in disparity space.
\begin{equation}
\Ldisparity\!(a, b) = \Big| \tfrac{1}{a_z} - \tfrac{1}{b_z}  \Big|_2^2\,.
\end{equation}
The disparity loss actually has the opposite bias of Euclidean loss:
it is minimized when scene scale grows very large (so that the disparities vanish).
This is not a problem for Luo et al., since they use fixed poses.
However, it affects our results negatively.

To alleviate this, we propose a new loss that measures the \emph{ratio} of depth values:
\begin{equation}
\Lratio\!(a, b) =
%\log \frac{\min(a_z, b_z)}{\max(a_z, b_z)}.
\frac{\max(a_z,~b_z)}{\min(a_z,~b_z)} - 1.
\end{equation}
This loss does not suffer from any depth bias; it does neither encourage growing nor shrinking the scene scale.
The measure $\Lratio$ is also used to compute the depth-similarity of samples in the depth filter in Eq.~\ref{eq:filter_weights}.

In summary, we define the reprojection similarity loss as follows:
\begin{equation}
\Ld\!(a, b) = \Lspatial\!(a, b) + \Lratio\!(a, b)
\end{equation}

\ignorethis{
\topic{Final objective}

The final objective is
\begin{equation}
\!\!\!\!\!\minimize
\bigg(
\sum_{(i, j) \in P\vphantom{\Mflow_\ij}}
~
\sum_{p \in \Mflow_\ij}
\Lrepro_\ij\!\p
\!\bigg)
+ \Ldeform + \Lfocal
\label{eq:final}
\end{equation}

We alternate between optimizing Eq.~\ref{eq:final} w.r.t.~pose and depth for a fixed number of 11 pose and 10 depth iterations (starting and ending with pose).
}

\ignorethis{
\topic{Negative results}
\johannes{
What about spatial alignment? It seemed useful on some sequences in the family videos?

Tried bilateral grids.

Tried bicubic instead of bilinear.}
}

%% file: figures/overview.tex
\begin{figure}%
\centering%
\includegraphics[width=\linewidth]{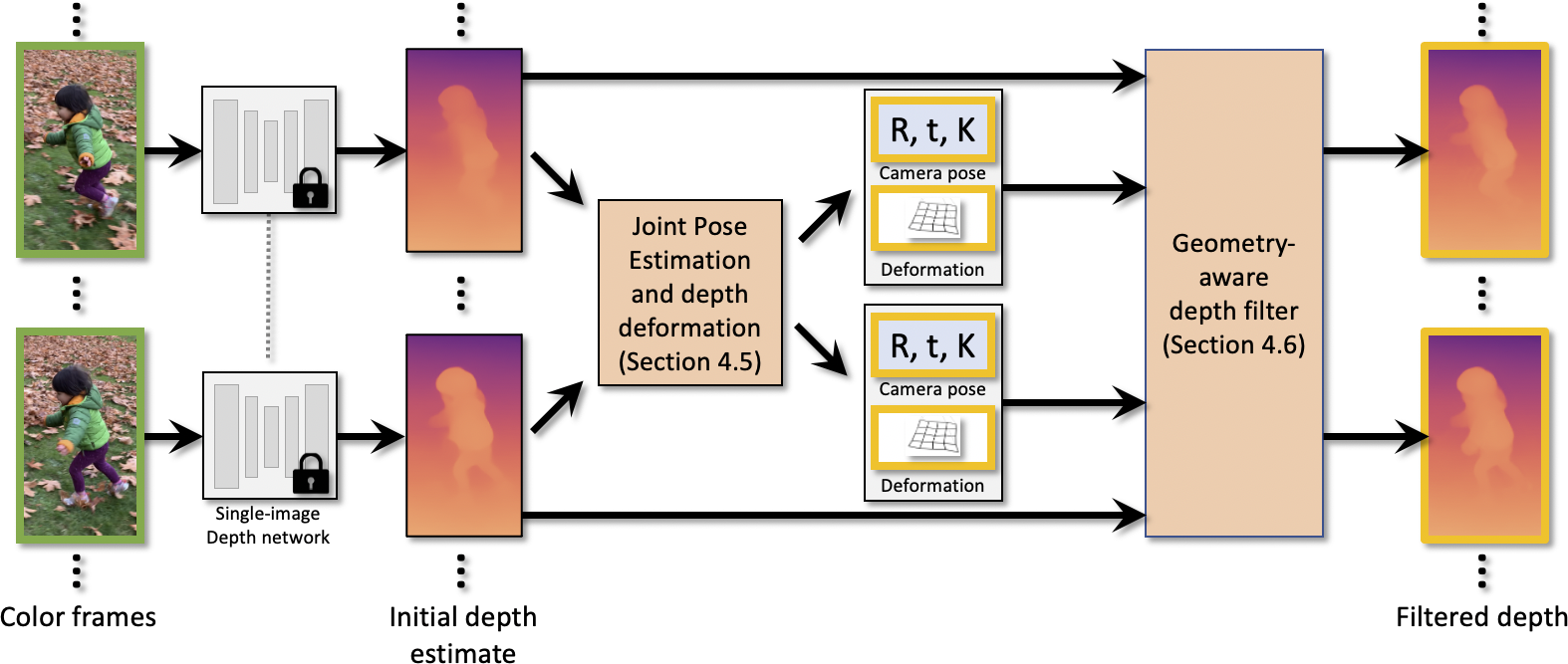}\\%
\caption{\textbf{Overview.}
Our algorithm only takes a monocular color video as input.
We first estimate per-frame depth maps using an existing single-frame CNN.
We jointly optimize camera poses as well as flexible deformations to align the depth maps in 3D and resolve any large-scale misalignments.
Finally, we resolve fine-scale details using a geometry-aware depth filter.
Green frames: inputs; yellow frames: outputs.
}%
\vspace{-1em}
\label{fig:overview}
\end{figure}%

%% file: figures/opt.tex
\providelength\width%
\setlength\width{2.43cm}%
\providelength\height%
\setlength\height{4cm}%
\providelength\gap%
\setlength\gap{0.724mm}
\providelength\bump%
\setlength\bump{0mm}%
\providelength\two%
\setlength\two{\dimexpr(\width+\gap+\width)}%
\providelength\three%
\setlength\three{\dimexpr(\two+\gap+\width)}%
\providelength\four%
\setlength\four{\dimexpr(\two+\gap+\two)}%

\newcommand{\cellleft}[1]{%
\begin{minipage}[t]{\width}%
\small #1%
\end{minipage}}%

\newcommand{\cellbig}[1]{%
\begin{minipage}[t]{\width}\centering%
#1%
\end{minipage}}%

\newcommand{\cell}[1]{%
\begin{minipage}[t]{\width}\centering%
\small #1%
\end{minipage}}%

\newcommand{\sepline}{%
\vspace{-2.8mm}\noindent\rule{\textwidth}{0.5pt}\\\vspace{-1.8mm}%
}

\newcommand{\cskip}{\hspace{4.4mm}}

\begin{figure*}%
\centering%	

\begin{minipage}[t]{\width}\centering%
\includegraphics[width=\width]{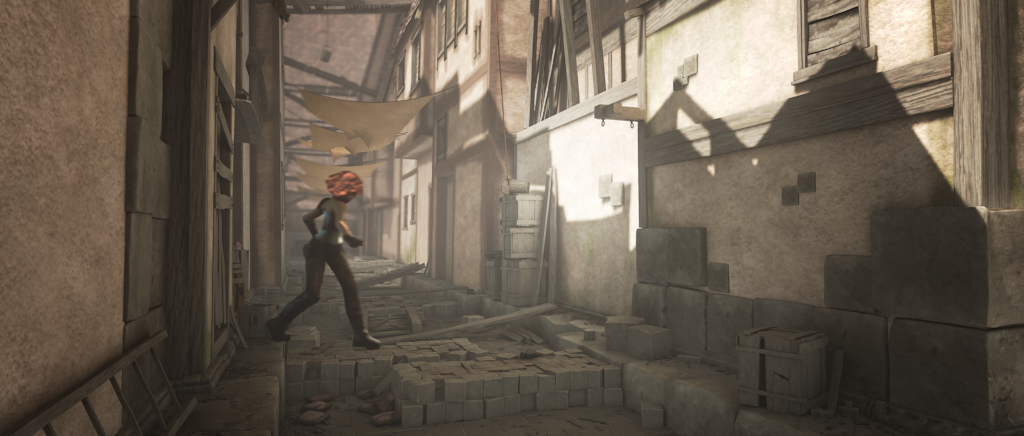}\\%
\includegraphics[width=\width,trim=250 70 630 140,clip]{figures/opt/color.png}\\%
\end{minipage}%
\hspace{\gap}%
\begin{minipage}[t]{\width}\centering%
\includegraphics[width=\width]{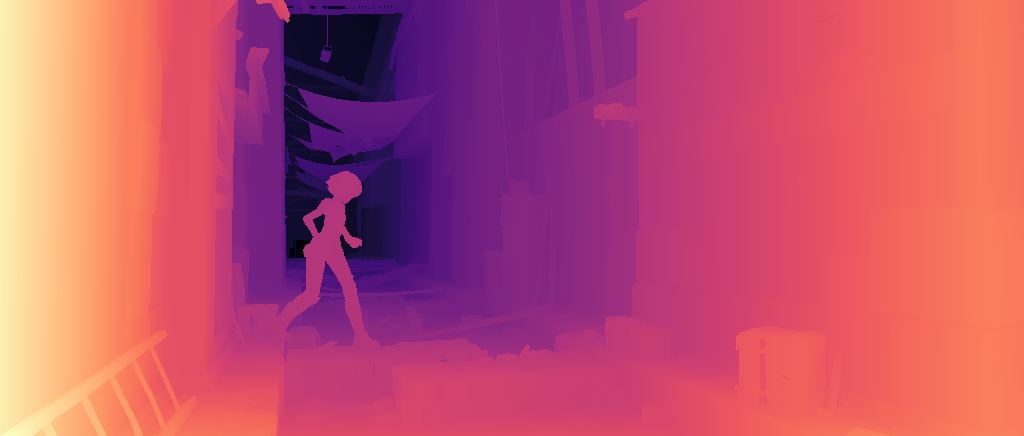}\\%
\includegraphics[width=\width,trim=250 70 630 140,clip]{figures/opt/GT_depth.png}\\%
\includegraphics[height=0.5\height]{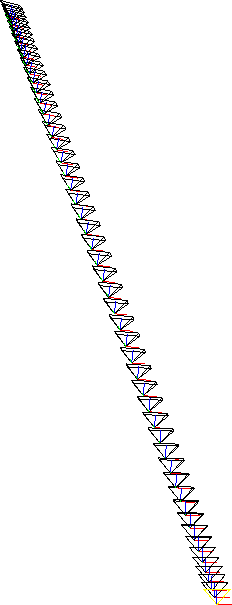}\\%
\end{minipage}%
\hspace{\gap}%
\begin{minipage}[t]{\width}\centering%
\includegraphics[width=\width]{figures/opt/GT_depth.png}\\%
\includegraphics[width=\width,trim=250 70 630 140,clip]{figures/opt/GT_depth.png}\\%
\includegraphics[height=0.5\height]{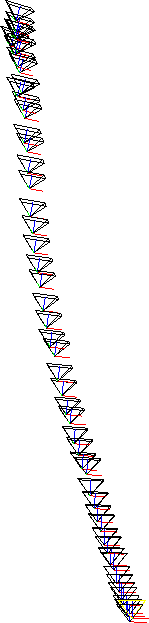}\\%
\end{minipage}%
\hspace{\gap}%
\begin{minipage}[t]{\width}\centering%
\includegraphics[width=\width]{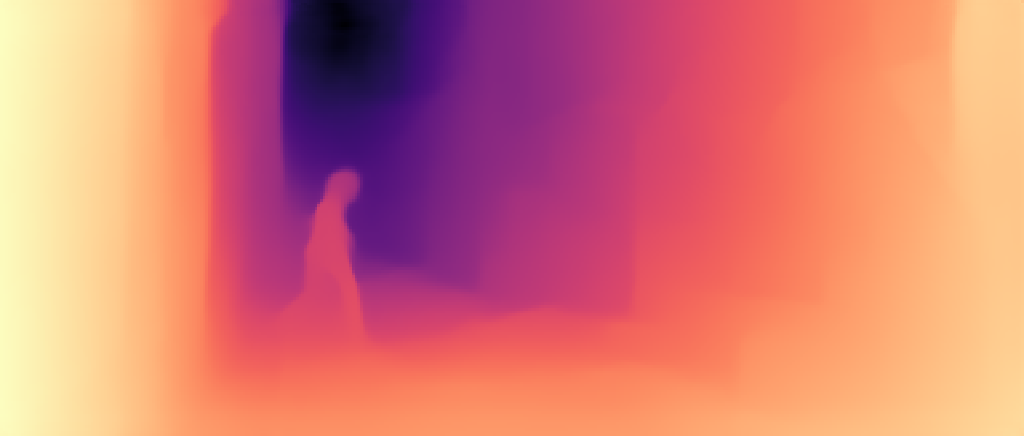}\\%
\includegraphics[width=\width,trim=250 70 630 140,clip]{figures/opt/EST_depth_SCALE_pose.png}\\%
\includegraphics[height=0.5\height]{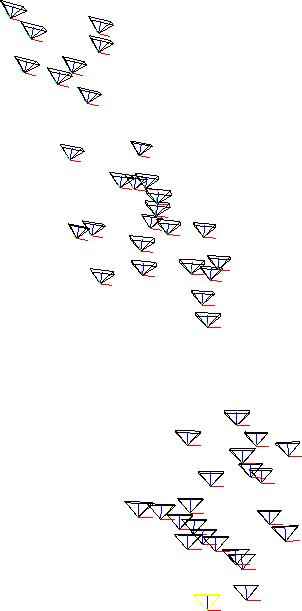}\\%
\end{minipage}%
\hspace{\gap}%
\begin{minipage}[t]{\width}\centering%
\includegraphics[width=\width]{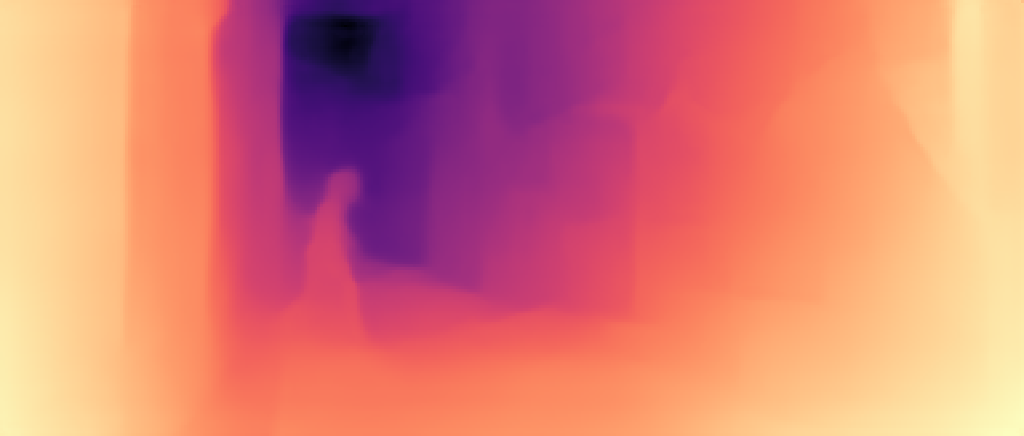}\\%
\includegraphics[width=\width,trim=250 70 630 140,clip]{figures/opt/EST_depth_SCALE_pose_FINETUNE.png}\\%
\includegraphics[height=0.5\height]{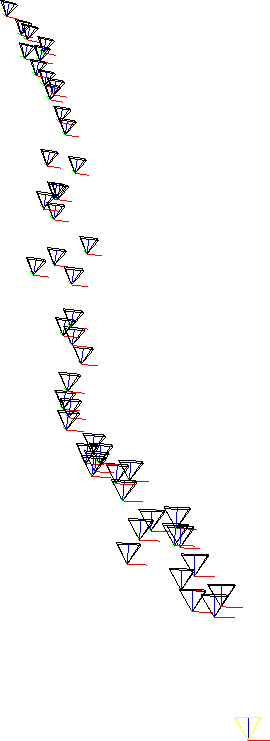}\\%
\end{minipage}%
\hspace{\gap}%
\begin{minipage}[t]{\width}\centering%
\includegraphics[width=\width]{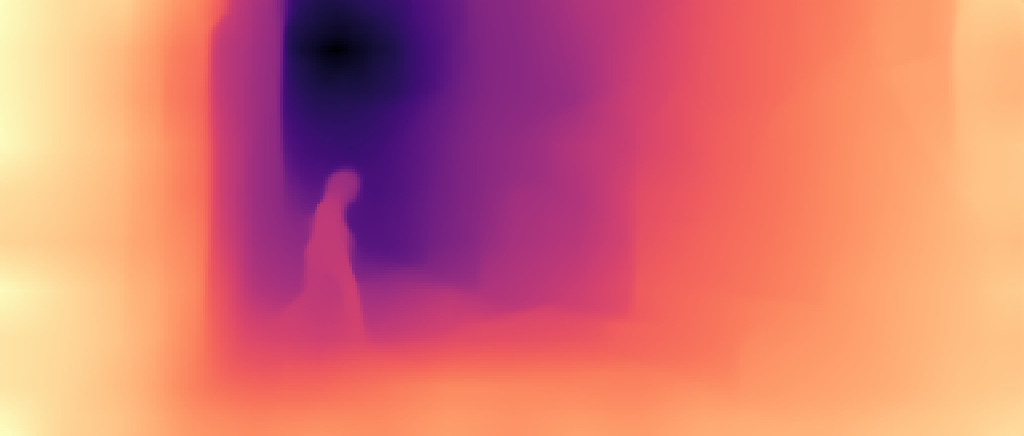}\\%
\includegraphics[width=\width,trim=250 70 630 140,clip]{figures/opt/EST_depth_FLEX_pose.png}\\%
\includegraphics[height=0.5\height]{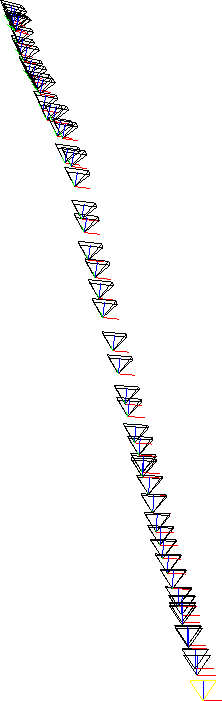}\\%
\end{minipage}%
\hspace{\gap}%
\begin{minipage}[t]{\width}\centering%
\includegraphics[width=\width]{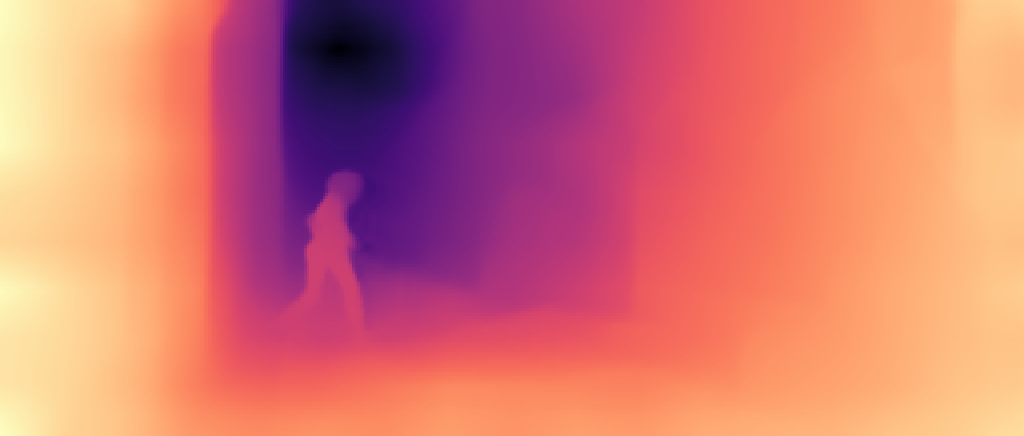}\\%
\includegraphics[width=\width,trim=250 70 630 140,clip]{figures/opt/EST_depth_FLEX_pose_FILTER.png}\\%
\includegraphics[height=0.5\height]{figures/opt/pose_flexible_initial_opt.png}\\%
\end{minipage}\\\vspace{0.5cm}%
\sepline%
\cellleft{Depth:}%
\hspace{\gap}%
\cellleft{\cskip Ground truth}%
\hspace{\gap}%
\cellleft{\cskip Ground truth}%
\hspace{\gap}%
\cellleft{\cskip Estimated \cite{lasinger2019towards}}%
\hspace{\gap}%
\cellleft{\cskip Estimated \cite{lasinger2019towards}}%
\hspace{\gap}%
\cellleft{\cskip Estimated \cite{lasinger2019towards}}%
\hspace{\gap}%
\cellleft{\cskip \bf Estimated \cite{lasinger2019towards}}\\%
\sepline%
\cellleft{Pose optimization:}%
\hspace{\gap}%
\cellleft{\cskip Ground truth}%
\hspace{\gap}%
\cellleft{\cskip Single scale ($s_i$)}%
\hspace{\gap}%
\cellleft{\cskip Single scale ($s_i$)}%
\hspace{\gap}%
\cellleft{\cskip Single scale ($s_i$)}%
\hspace{\gap}%
\cellleft{\cskip Flexible ($\varphi_i$)}%
\hspace{\gap}%
\cellleft{\cskip \bf Flexible ($\varphi_i$)}\\%
\sepline%
\cellleft{Depth refinement:}%
\hspace{\gap}%
\cellleft{\cskip \textcolor{gray}{N/A}}%
\hspace{\gap}%
\cellleft{\cskip \textcolor{gray}{N/A}}%
\hspace{\gap}%
\cellleft{\cskip \textcolor{gray}{N/A}}%
\hspace{\gap}%
\cellleft{\cskip Fine-tuning}%
\hspace{\gap}%
\cellleft{\cskip \textcolor{gray}{N/A}}%
\hspace{\gap}%
\cellleft{\cskip \bf Depth filter}\\%
\sepline\vspace{0.5mm}%
\cellleft{~}%
\hspace{\gap}%
\cellbig{(a)}%
\hspace{\gap}%
\cellbig{(b)}%
\hspace{\gap}%
\cellbig{(c)}%
\hspace{\gap}%
\cellbig{(d)}%
\hspace{\gap}%
\cellbig{(e)}%
\hspace{\gap}%
\cellbig{\bf (f)}%
\vspace{-1mm}%
\caption{\textbf{Joint depth and pose optimization.} Various configurations of our algorithm: (a-b) Ground truth depth with ground truth and estimated poses, respectively. (c) Misalignments in estimated depth impose jittery errors on the optimized camera trajectories. (d) CVD-style fine-tuning fails in the absence of precise poses. (e) Our flexible deformations resolve depth misalignments, which results in smoother camera trajectories. (f) Using geometry-aware depth filtering we resolve fine depth details (our final result).}%
\vspace{-2mm}
\label{fig:opt}
\end{figure*}

%% file: figures/filter.tex
\providelength\width%
\setlength\width{1.95cm}
\begin{figure}%
\centering%
\rotatebox{90}{\hspace{1cm}Input}%
\hfill%
\includegraphics[width=\width]{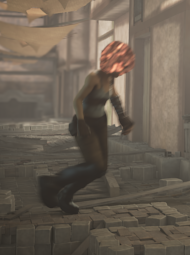}%
\hfill%
\includegraphics[width=\width]{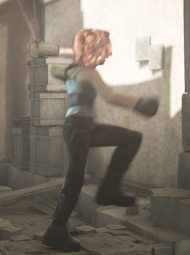}%
\hfill%
\includegraphics[width=\width]{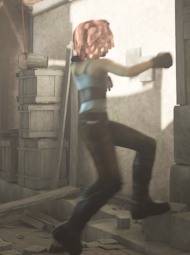}%
\hfill%
\includegraphics[width=\width]{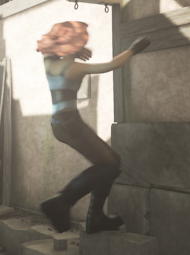}%
\\%
\rotatebox{90}{\hspace{0.4cm}Initial depth}%
\hfill%
\includegraphics[width=\width]{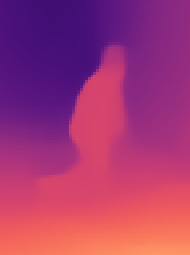}%
\hfill%
\includegraphics[width=\width]{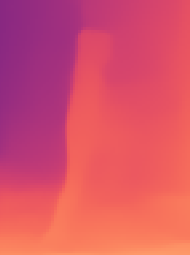}%
\hfill%
\includegraphics[width=\width]{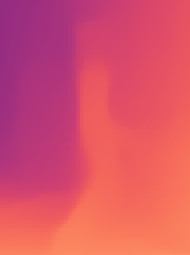}%
\hfill%
\includegraphics[width=\width]{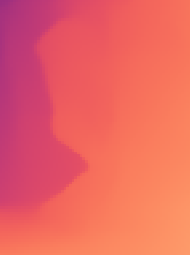}%
\\%
\rotatebox{90}{\hspace{0.3cm}Filtered depth}%
\hfill%
\includegraphics[width=\width]{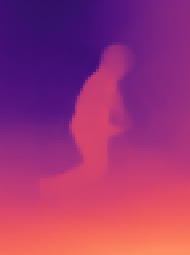}%
\hfill%
\includegraphics[width=\width]{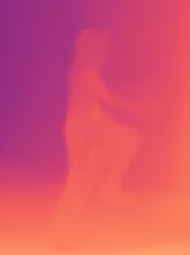}%
\hfill%
\includegraphics[width=\width]{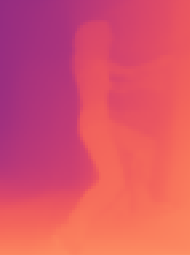}%
\hfill%
\includegraphics[width=\width]{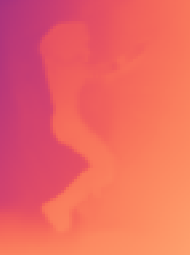}%
\\\vspace{-1mm}%
\caption{
\textbf{Geometry-aware depth filter.}
Top: frames from the input video.
Middle: initial per-frame depth estimates (after flexible alignment).
Bottom: the geometry-aware depth filter resolves fine-scale details in the depth maps.
}
\vspace{-4mm}
\label{fig:filter}
\end{figure}%

%% file: 4_result.tex
\input{table/sintel_eval}

\section{Experimental Results}
\label{sec:results}

%\input{figures/strip.tex}

%\johannes{Make a point about how flexible deformation improves the pose estimation. The black swan sequence is a good example. There are many others. We can show this visually in a figure, and also quantitatively.} 

\subsection{Experimental setup}
\topic{Datasets.}
We validate the effectiveness of the proposed method on three main video datasets, covering a wide range of challenging indoor and outdoor scenes. 
\changed{Here we focus on reporting results on the Sintel~\cite{ButlerSintel2012} dataset.
The MPI Sintel dataset consists of 23 synthetic sequences of highly dynamic scenes. 
Each sequence comes with ground truth depth measured in meters as well as ground truth camera poses. 
We use both the \emph{clean} and \emph{final} versions of the dataset.
The ground truth annotations allow us to conduct a quantitative comparison on both the estimated depth and pose.
We refer the readers to our qualitative results on \tb{DAVIS~\cite{perazzi2016benchmark}} and \tb{Cellphone videos} in the project page.
}

% \begin{itemize}[leftmargin=0.2cm]
% \item \tb{Sintel~\cite{ButlerSintel2012}}: The MPI Sintel dataset consists of 23 synthetic sequences of highly dynamic scenes. 
% Each sequence comes with ground truth depth measured in meters as well as ground truth camera poses. 
% We use both the \emph{clean} and \emph{final} versions of the dataset.
% The ground truth annotations allow us to conduct a quantitative comparison on both the estimated depth and pose.
% \item \tb{DAVIS~\cite{perazzi2016benchmark}}: The DAVIS dataset contains 90 videos (trainval set in the 2017 DAVIS challenge). 
% The dataset was originally designed for evaluating video object segmentation algorithms. 
% The camera motions in many videos are rotation with limited translation.
% Consequently, the lack of reliable disparity cues renders it particularly challenging for conventional multi-view stereo algorithms.
% \item \tb{\tb{Cellphone videos} }: 
% To further evaluate our proposed method's robustness, we collect 71 casually captured videos from a hand-held cell phone camera.
% These videos present additional challenges that are not covered by existing carefully curated datasets, such as higher-level camera noise, motion blur, non-rigid dynamic objects, and complex background.
% As we do not have ground truth depth or camera poses for such in the wild videos. 
% We show visual results of the recovered camera trajectories and the depth. 
% \end{itemize}
Note that we do not choose datasets focusing on \emph{closed-domain applications} such as driving scenes (e.g., KITTI depth/odometry dataset or office (e.g., TUM RGB-D~\cite{Sturm-TUMRGBD2012}) or \emph{fully static} scenes (e.g., ScanNet~\cite{dai2017scannet}). 
% Methods specifically designed for those applications may work better.

\topic{Compared methods.}
We compare our results with several state-of-the-art depth estimation algorithms.
\begin{itemize}[leftmargin=0.2cm]
\setlength\itemsep{0.05em}
\item \tb{COLMAP~\cite{schonberger2016structure}}: Traditional SFM/MVS algorithm for 3D reconstruction. As COLMAP reconstruction is very sensitive to dynamic objects in the scene, we use the same $\Mdyn$ dynamic masks using by our algorithm (computed automatically Mask R-CNN) to exclude feature extraction/matching from those areas. 
%\jiabin{Fill in specific setting for running COLMAP?}
\item \tb{DeepV2D~\cite{teed2020deepv2d}}: Video depth estimation algorithm using differentable motion estimation and depth estimation. 
\item \tb{CVD~\cite{Luo-VideoDepth-2020}}: A hybrid method that combine multi-view and single-view depth estimation methods for producing consistent video depth. 
\item \tb{MiDaS-v2~\cite{lasinger2019towards}}: State-of-the-art single-view depth estimation model. 
\end{itemize}

% \topic{Pose methods}
% - D3VO: Deep Depth, Deep Pose and Deep Uncertainty for Monocular Visual Odometry \cite{yang2020d3vo} (no code)
% - ORB-SLAM2
% - COLMAP

\subsection{Evaluation on MPI Sintel dataset~\cite{ButlerSintel2012}}
% Dataset information
% The depth values are returned from Blender in an additional Z-buffer pass, similar to the optical flow.
% The extrinsic camera matrix is given by Blender, as the world matrix of the camera object.
% The intrinsic camera matrix is computed using the focal length as given by Blender, the hard-coded pixel dimensions of 32 px / mm, zero pixel skew, and the principal point at x = 511.5 and y = 217.5.

% \input{table/sintel_depth}

\topic{Depth evaluation.}
As depth estimation from all the methods is up to an unknown scene scale, we follow the standard depth evaluation protocol to align the predicted depth and the ground truth depth using median scaling. 
We exclude depth values that are larger than 80 meters.
\tabref{sintel_depth} shows the quantitative comparisons with the state-of-the-art on various metrics. 
%\jiabin{Describe how awesome our method is.}
% We use the 22 of the 23 

Note that COLMAP~\cite{schonberger2016structure} fails to estimate the camera pose for 11 in 23 sequences.
CVD~\cite{Luo-VideoDepth-2020}, which uses COLMAP poses as input, thus, also does not produce depth estimation results for these scenes. 
Furthermore, COLMAP discards single-pixel depth estimates deemed unreliable.
As a result, we cannot evaluate these methods using standard \emph{averaged} metrics. 
Instead, we store \emph{all} the pixel-wise error metrics across all pixels, frames, and videos. 
We then sort these errors and plot the curves. 
Please refer to the supplemental material.%\figref{sintel_depth_partial}. 
The plots capture both the \emph{accuracy} and \emph{completeness} of each method. 

%\input{figures/sintel_depth_partial}

% Depth evaluation
% - [Table] Per-frame alignment (scale-invariant metrics)
% - [Table] Per-sequence alignment (global scale)

\topic{Pose evaluation.}
We follow the standard evaluation protocol of visual odometry for pose evaluation and compare our methods against state-of-the-arts in terms of absolute trajectory error (ATE) and relative pose error (RPE). ATE measures the root-mean-square error between predicted camera poses $[x, y, z]$ and ground truth. RPE measures frame-to-frame relative pose error between frame pairs, including translation error (RPE-T) and rotational error (RPE-R). Still, since the scene scale is unknown, we scale and align the predictions to the ground truth associated poses during evaluation by minimizing ATE for fair comparisons.

We conduct pose evaluations on Sintel with ground truth pose annotations, and the quantitative results are presented in Table~\ref{tab:sintel_eval} and Figure~\ref{fig:rpe}. 
As demonstrated in Table~\ref{tab:sintel_eval}, our proposed method outperforms MiDaS-v2 on all metrics (mean of ATE, RPE-T, and RPE-R) with a noticeable margin and significantly outperforms DeepV2D, on both the Clean and Final categories of Sintel.
Note that since COLMAP and CVD fail on a significant portion of the dynamic scenes of Sintel, it is not directly comparable in terms of mean errors as in Table~\ref{tab:sintel_eval}. 
For a fair comparison, we store RPE-T and RPE-R errors from all plausible pose predictions between frame pairs from all Sintel sequences, then sort them and plot the curves in Figure~\ref{fig:rpe}.
The results show our proposed method consistently achieves more accurate pose predictions than completing methods.

% Pose evaluation
% \input{table/sintel_pose}
\input{figures/rpe}

\subsection{\changed{Limitations}}

\changed{
The main limitation that we observe is a kind of residual wobble of the aligned depth maps.
It is apparent in most results we provide on the project page.
We think that it can be addressed by better deformation models, in particular, replacing the spline-based deformation with a pixel-based deformation field with appropriate regularization.
However, this would require denser pairwise constraints, which our current formulation using continuous global optimization does not support, i.e., when we densify the constraints the Ceres Solver memory usage blows up and the performance goes down drastically.
This is due to the global nature of the optimization problem.
Finding a better formulation to resolve this problem is a great avenue for future work.}

%% file: table/sintel_eval.tex
\begin{table*}[t]
\centering
\caption{
\textbf{Quantitative evaluations of depth and pose on the MPI Sintel benchmark} (Top: \textit{Sintel Clean}, Bottom: \textit{Sintel Final}). For depth evaluation, we present per-frame evaluations on standard error and accuracy metrics. For pose evaluation, we present per-sequence evaluations on translational and rotational error metrics. 
}
\vspace{-4mm}
\label{tab:sintel_depth}
\mpage{0.98}{
\small \vspace{1em}
\resizebox{\linewidth}{!}{
\renewcommand{\arraystretch}{1.2} % Default value: 1
\begin{tabular}{lccccccccccccc}
\toprule
& \multicolumn{4}{c}{\texttt{Depth - Error metric} $\downarrow$} && \multicolumn{3}{c}{\texttt{Depth - Accuracy metric} $\uparrow$} &&
\multicolumn{3}{c}{\texttt{Pose - Error metric} $\downarrow$}\\
\cmidrule{2-5} \cmidrule{7-9} \cmidrule{11-13}
Method  &  \textbf{Abs Rel}  & \textbf{Sq Rel}  & \textbf{RMSE}  & \textbf{log RMSE}  && $\mathbf{\delta < 1.25}$ & $\mathbf{\delta < 1.25^2}$ & $\mathbf{\delta < 1.25^3}$ && \textbf{ATE (m)$\downarrow$} & \textbf{RPE Trans (m)$\downarrow$} & \textbf{RPE Rot (deg)$\downarrow$}\\
\midrule
DeepV2D~\cite{teed2020deepv2d} &   0.526  &   3.629  &   6.493  &   0.683  &&   0.487  &   0.671  &   0.761 && 0.9526 & 0.3819 & 0.1869 \\
Ours - Single-scale pose (aligned MiDaS) &   0.380  &   \ul{2.617}  &   \ul{5.773}  &   0.533  &&   0.562  &   0.736  &   0.832  && 0.1883 & 0.0806 & 0.0262 \\
Ours - Single-scale pose + depth fine-tuning &   0.472  &   3.444  &   6.340  &   0.635  &&  0.534  &  0.694  &   0.790  && \ul{0.1686}  &  0.0724  & 0.0139 \\
Ours - Single-scale pose + depth filter &   \textbf{0.375}  & \textbf{2.546}  &   \textbf{5.763}  &   \textbf{0.530}  &&  \textbf{0.569}  &  0.738  &   0.835  && 0.1882 &  0.0806  & 0.0262 \\
Ours - Flexible pose &   0.379  &   2.702  &   5.795  &   0.533  &&  0.565  &  \ul{0.744}  &   \ul{0.836}  && 0.1843  &  \ul{0.0723}  & \ul{0.0095} \\
Ours - Flexible pose + depth fine-tuning &   0.439  &   3.100  &   6.213  &   0.614 &&  0.524  &  0.698  &   0.796  && \textbf{0.1656}  & \textbf{0.0651}  & \textbf{0.0070} \\
\textbf{Ours - Flexible pose + depth filter} &   \ul{0.377}  &   2.657  &   5.786  &   \ul{0.531}  &&   \ul{0.568}  &  \textbf{0.745}  &   \textbf{0.837}  && {0.1843}  &  \ul{0.0723}  & \ul{0.0095}\\

\midrule
 DeepV2D &   0.526  &   3.620  &   6.470  &   0.670  &&   0.486  &   0.674  &   0.760  && 0.9192 & 0.5834 & 0.2506 \\
Ours - Single-scale pose (aligned MiDaS) &  0.425  &   2.640  &   \ul{5.858}  &   0.559  &&   \ul{0.529}  &   0.726  &   0.828  && 0.2210  &  0.0827  &  0.0258\\
Ours - Single-scale pose + depth fine-tuning &   0.473  &   3.215  &   6.298  &   0.639  &&  0.527  &  0.684 &   0.782  && \ul{0.1620}  &  0.0727  & 0.0116 \\
Ours - Single-scale pose + depth filter &   \ul{0.421}  &   \textbf{2.616}  &   \textbf{5.850}  &   \textbf{0.556}  &&  \textbf{0.533}  &  \ul{0.728}  &   0.830  && 0.1803  &  0.0827  & 0.0258 \\
Ours - Flexible pose &   \ul{0.421}  &   2.660  &   5.906  &   0.559  &&  0.523  &  \textbf{0.730}  &   \ul{0.832}  && 0.1831  &  \ul{0.0713}  & \ul{0.0088} \\
Ours - Flexible pose + depth fine-tuning &   0.438  &   3.053  &   6.300  &   0.605  &&  0.525  &  0.705  &   0.807  && \textbf{0.1594}  &  \textbf{0.0652}  & \textbf{0.0073} \\
\textbf{Ours - Flexible pose + depth filter} &   \textbf{0.419}  &   \ul{2.628}  &  5.896  &  \ul{0.558}  &&   0.526  &   \textbf{0.730}  &   \textbf{0.833} && {0.1831}  &  {0.0714} &  \ul{0.0088} \\
\bottomrule
\end{tabular}
}
}
\vspace{-4mm}
\label{tab:sintel_eval}
\end{table*}

%% file: figures/rpe.tex
\begin{figure}[t!]
\centering
\captionsetup[subfigure]{labelformat=empty}

% \par\medskip

\begin{subfigure}{.48\linewidth}
\centering
\captionsetup{width=0.98\linewidth}
\includegraphics[width=0.98\linewidth]{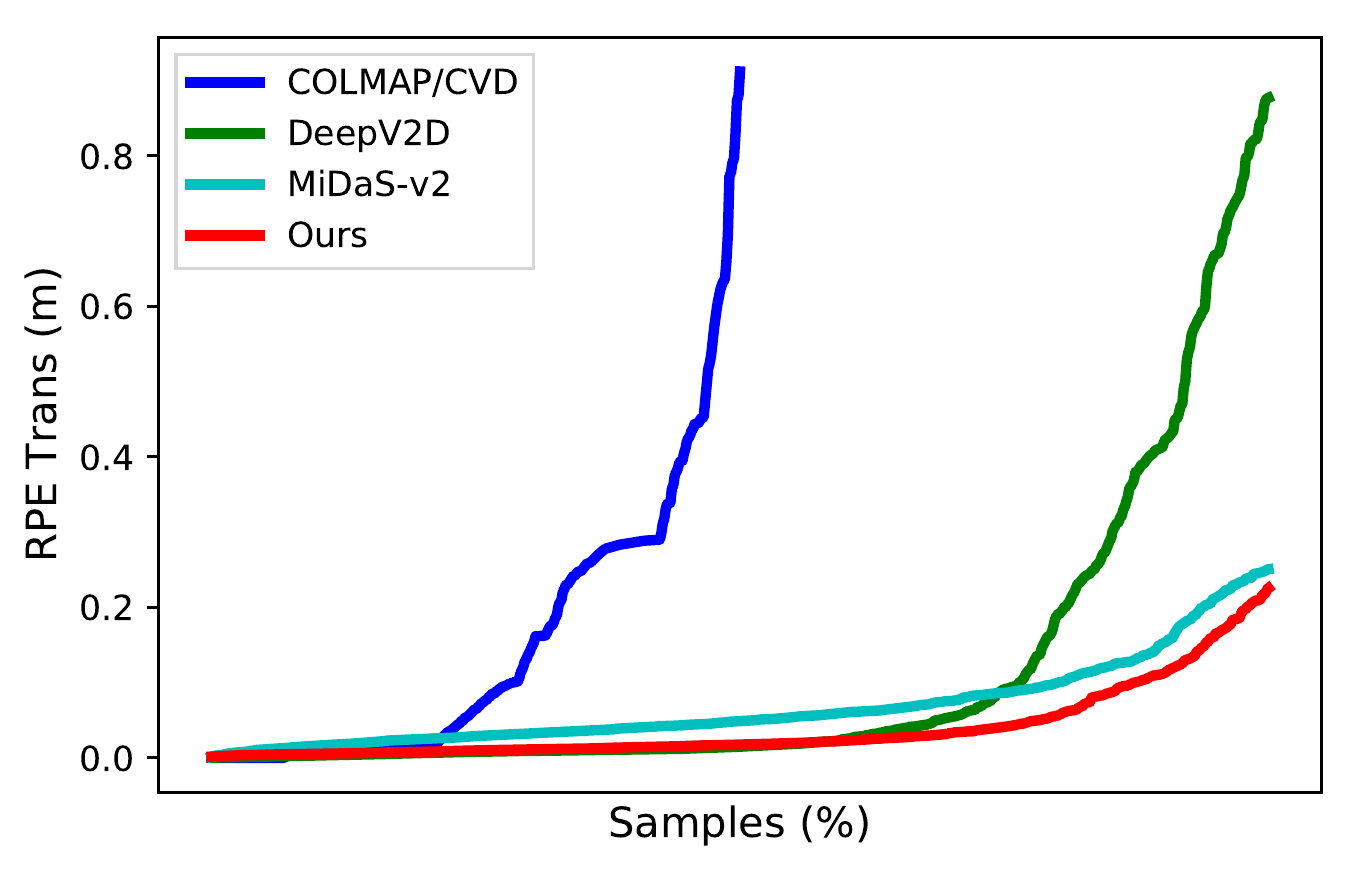}\vspace{-2.5mm}
\caption{{\footnotesize (a) RPE Trans - Sintel Clean}}
\end{subfigure}
\begin{subfigure}{.48\linewidth}
\centering
\captionsetup{width=0.98\linewidth}
\includegraphics[width=0.98\linewidth]{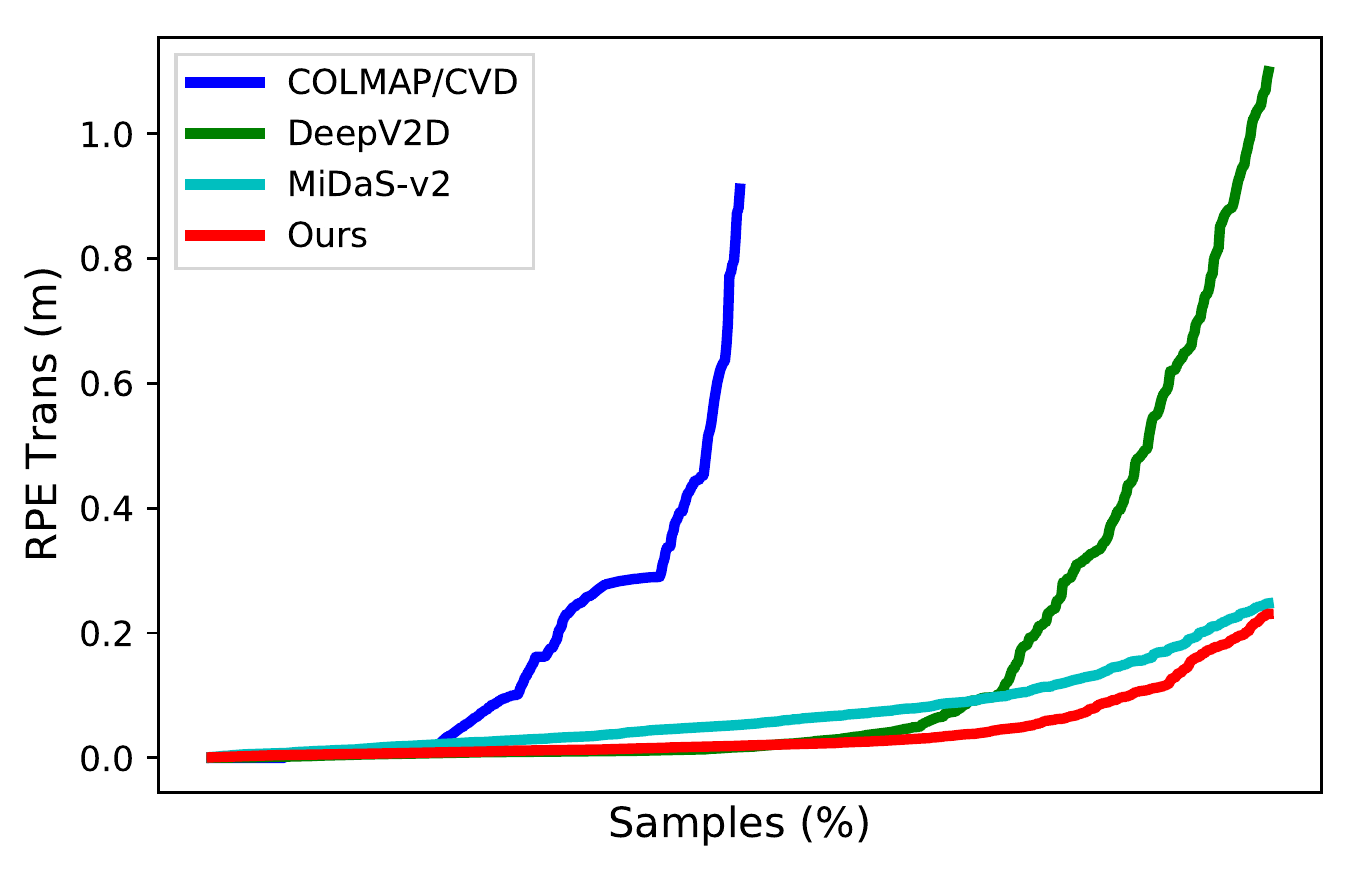}\vspace{-2.5mm}
\caption{{\footnotesize (b) RPE Trans - Sintel Final}}
\end{subfigure}
\medskip
\begin{subfigure}{.48\linewidth}
\centering
\captionsetup{width=0.98\linewidth}
\includegraphics[width=0.98\linewidth]{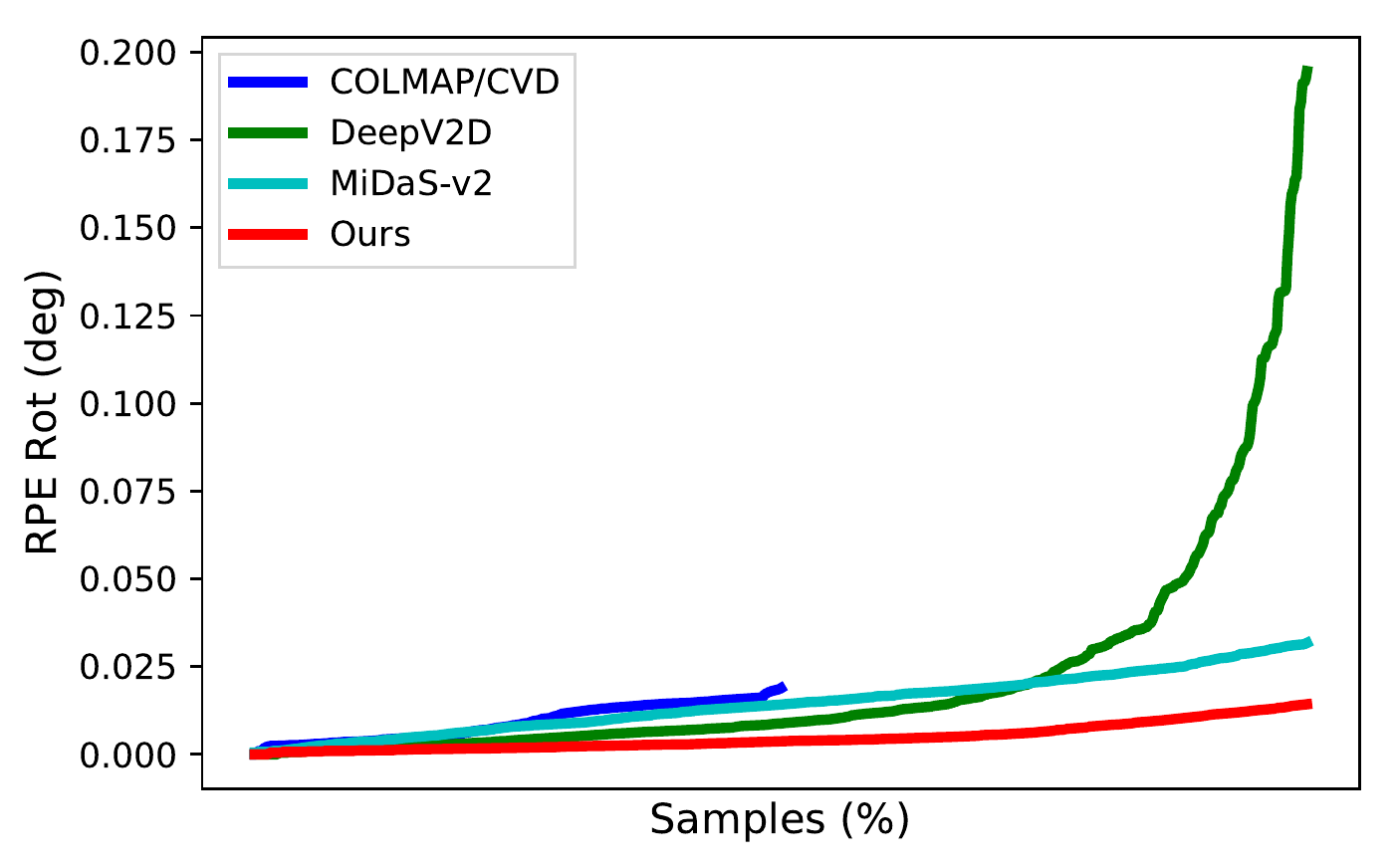}\vspace{-2.5mm}
\caption{{\footnotesize (c) RPE Rot - Sintel Clean}}
\end{subfigure}
\begin{subfigure}{.48\linewidth}
\centering
\captionsetup{width=0.98\linewidth}
\includegraphics[width=0.98\linewidth]{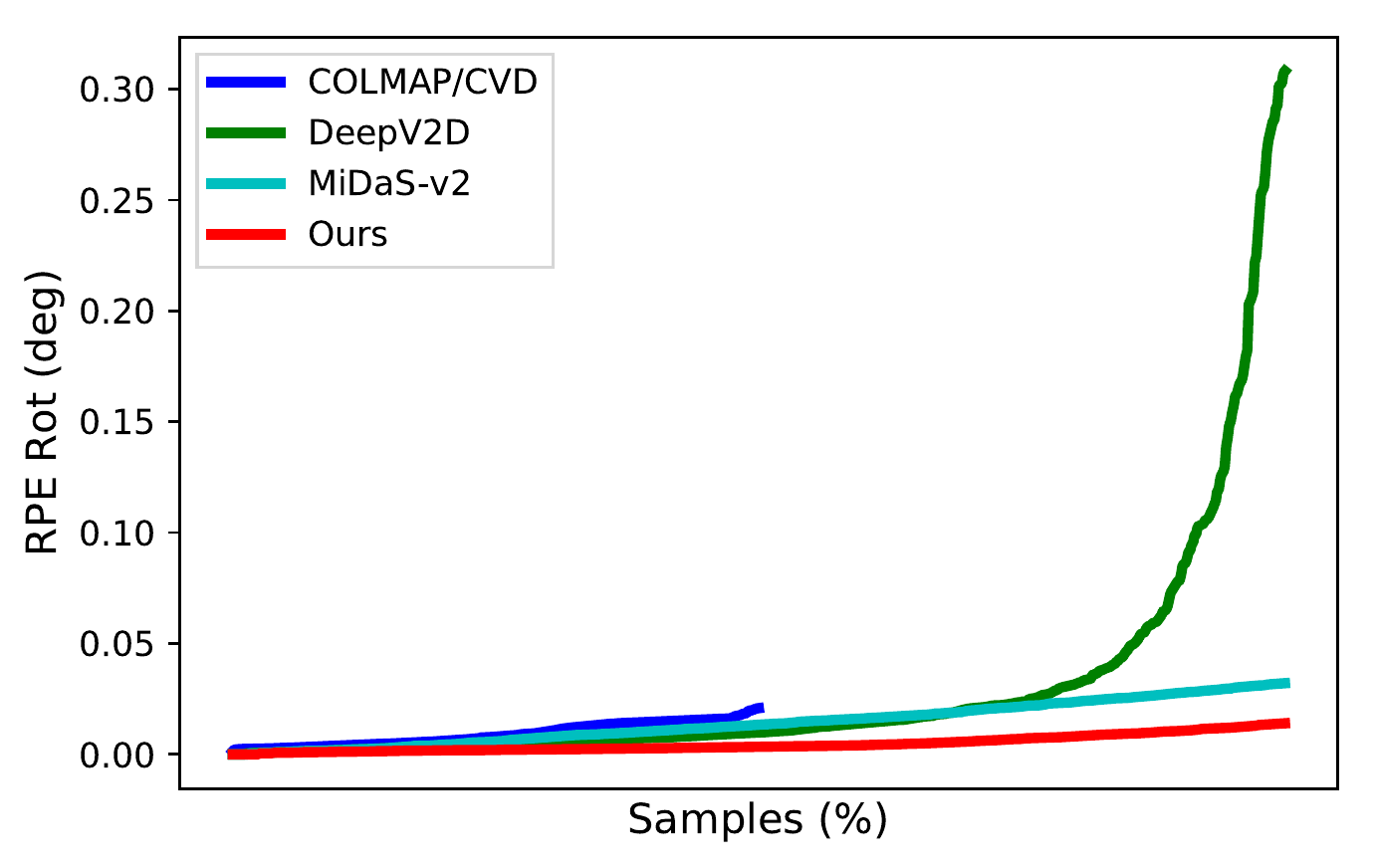}\vspace{-2.5mm}
\caption{{\footnotesize (d) RPE Rot - Sintel Final}}
\end{subfigure}
\vspace{-4mm}
\caption{\small \tb{Evaluation of translational and rotational Relative Pose Error (RPE) on Sintel}. All the frame pair-wise errors are stored and sorted for plotting the distributions. Note that COLMAP and CVD (which relies on COLMAP) fail on many Sintel sequences, resulting in partial data points.}
\vspace{-4mm}
\label{fig:rpe}
\end{figure}

%% file: 5_conclusion.tex
\section{Conclusions}
\label{sec:conclusions}

We presented a general optimization algorithm for consistent depth estimation on monocular videos, requiring neither input poses nor inference-time fine-tuning. 
Our method attains robust reconstruction for challenging dynamic videos casually captured by hand-held devices, and achieves better performances on diverse test beds.

%\xuejian{Then one or two sentences for future work?}
\newpage

%% file: sup_content.tex
In this supplementary document, we present additional implementation details, quantitative evaluation, and visual results to complement the main paper \cite{kopf2021robust}

\section{Visual Results of Depth and Pose Estimation}
We provide the visual comparisons with the state-of-the-art depth estimation algorithms in the accompanying videos.
Specifically, we show the following three videos:
\begin{enumerate}
\item \tb{Cellphone videos}: 
We show visual comparisons of our results with Consistent Video Depth (CVD)~\cite{Luo-VideoDepth-2020}, COLMAP~\cite{schonberger2016structure}, and DeepV2D~\cite{teed2020deepv2d}.
\item \tb{Sintel dataset}: We use sample videos from the Sintel dataset to validate the two core technical contributions of our work (1) flexible depth deformation for pose optimization and (2) geometry-aware depth filtering.
\item \tb{DAVIS dataset}: To demonstrate the robustness of our approach, we present the results of \emph{all} 90 videos from the DAVIS 2017 train and validation dataset.
\end{enumerate}

We will release the source code and the results (dense depth maps and camera poses).

\section{Runtime Analysis}

% FIGURE 1 (Runtime vs video length)
% - X-axis: video length (50, 100, 200, 400, 800)
% - Y-axis: time spent

% FIGURE 2 (Profiling)
% - Pick a video length, e.g., 100
% - Calculate the percentage of the time spent on 
% a) depth estimation (MiDaS-v2)
% b) Flow estimation (RAFT)
% c) Pose estimation + depth deformation
% d) depth filter

We provide a runtime analysis and profiling of our method. 
Given a long input video, we test our method with a varying length of input video frames (first $K$ frames from the video).
The original video resolution: $1920\times1080$; downscaled resolution for optical flow estimation~\cite{Teed2020}: $1024\times576$; downscaled resolution for initial depth estimation~\cite{lasinger2019towards}: $384\times224$). 
We use two NVIDIA Tesla V100 GPUs. 
Note that our depth filtering code is \emph{single-threaded} and would significantly benefit from multi-threading in practice. Figure~\ref{fig:ablation} and Table~\ref{tab:ablation} demonstrate the detailed analysis.

\topic{\changed{Discussion.}}
\changed{
The performance of our method is far from real-time, but compares favorably to many of the baseline methods we have tested:
COLMAP-dense~\cite{schoenberger2016mvs} runtime is very variable, but generally considerably slower than our method, on the order of hours or even days in our experiments.
CVD~\cite{Luo-VideoDepth-2020} typically runs for multiple hours on a short video clip.
DeepV2D~\cite{teed2020deepv2d} is faster (e.g. a few minutes for a 50-frame sequence); however, the quality is considerably lower than ours, and it quickly runs out of memory for longer sequences (the maximum were 75 frames on a NVIDIA 2080Ti GPU with 11GB memory).
We are not aware of any real-time dense video depth estimation methods that provide comparable quality to ours.
Most SLAM methods, for example, only provide sparse or semi-dense depth.
}

\input{table/ablation_runtime}

\input{figures/fig_runtime_analysis}

\section{\changed{Evaluation on ScanNet}}

\input{table/scannet}

\changed{Table~\ref{tab:scannet} provides a quantitative evaluation on the ScanNet dataset.}

\section{Additional Quantitative Evaluation on the Sintel Dataset}
% Sintel Depth-pose evaluation at sequence level
% Similar to Table 1, but for each video (just final)

\topic{Depth evaluation partial results.} 
\figref{sintel_depth_partial} shows a quantitative comparison with methods that produce \emph{partial} depth maps (i.e., depth is not reported for all pixels; e.g., COLMAP) or do not succeed in some of the sequences. 
To quantitatively evaluate the performance for methods with partial results, we store all the pixel-wise error metrics across all the pixels, frames, and videos and plot the sorted values. 
The plot here capture both the accuracy as well as the completeness of each method.

\input{figures/sintel_depth_partial}

\topic{Per-sequence evaluation.} 
We provide per-sequence quantitative evaluation for depth and pose on the Sintel dataset in \tabref{sintel_eval_seq_1}, \tabref{sintel_eval_seq_2}, and \tabref{sintel_eval_seq_3}. 

\input{table/sintel_eval_seq}

\section{\changed{Additional Discussion (from Rebuttal)}}

\topic{\changed{Can the inaccurate depth itself be directly optimized, instead of introducing an extra deformation model? What are the advantages of using a deformation model?}}
\changed{
(1) It will be too unconstrained.
In CVD, with fixed camera poses, optimizing depth values directly would reduce the problem to na{\"i}ve multi-view stereo triangulation (which would not produce good results).
In our case, where camera poses are part of the optimization variables, direct depth value optimization would not converge to a good solution, since the problem is non-linear and sensitive to initialization.
Overcoming the sensitivity to initialization would require good regularization.
CVD achieves it by fine-tuning CNN weights shared for all frames, rather than optimizing the depth values themselves.
In our case, we achieve the regularization by using smooth depth deformation functions, which enforce a similarity (up to scale) to the original depth produced by the pretrained depth network.
(2) The global optimization (for all frames) problem is too large.
The smooth deformation functions have fewer parameters, which makes global optimization feasible
}

\topic{\changed{Is smooth depth deformation an ideal solution?}}
\changed{
Ideally we would use a \emph{piece-wise} smooth deformation that is controlled at the pixel level.
However, we are currently bound by the memory and compute demands of the global deformation optimization.
Please refer to our response to the question about limitations above.
We believe this is a great avenue for future work.
}

%% file: table/ablation_runtime.tex
\begin{table}[ht!]
	\centering
	\caption{Runtime analysis of videos of varying length, broken down by algorithm stages. Times are reported in seconds.
	}
	\begin{minipage}[b]{0.98\linewidth}
		\centering
        \small
        \renewcommand{\arraystretch}{1.5} % Default value: 1
        \resizebox{\linewidth}{!}{
		\begin{tabular}{@{}l ccccc}
			\midrule[0.8pt]
			\textbf{Video Length in Frames} & \textbf{Optical Flow~\cite{Teed2020}} &  \textbf{Depth Estimation~\cite{lasinger2019towards}} & \textbf{Pose Optimization} & \textbf{Geometry-aware Depth Filtering} & \textbf{Total} \\ \midrule[0.8pt]
			$50$  & ~~252.0s & ~9.2s & ~36.6s & ~75.8s & ~~373.6s \\
			$100$  & ~~515.7s & 16.0s & ~54.5s & 172.0s & ~~758.3s\\
			$200$ & 1,268.9s & 30.9s  & ~94.8s & 334.9s & 1,729.5s\\
			$400$  & 2,611.4s & 58.3s & 175.2s & 704.1s & 3,549.0s\\
			\bottomrule[0.8pt]
		\end{tabular}
		}
	\end{minipage}
	\label{tab:ablation}
    % \vspace{-2em}
\end{table}

%% file: figures/fig_runtime_analysis.tex
\begin{figure}%
\centering%
\includegraphics[width=0.6\linewidth]{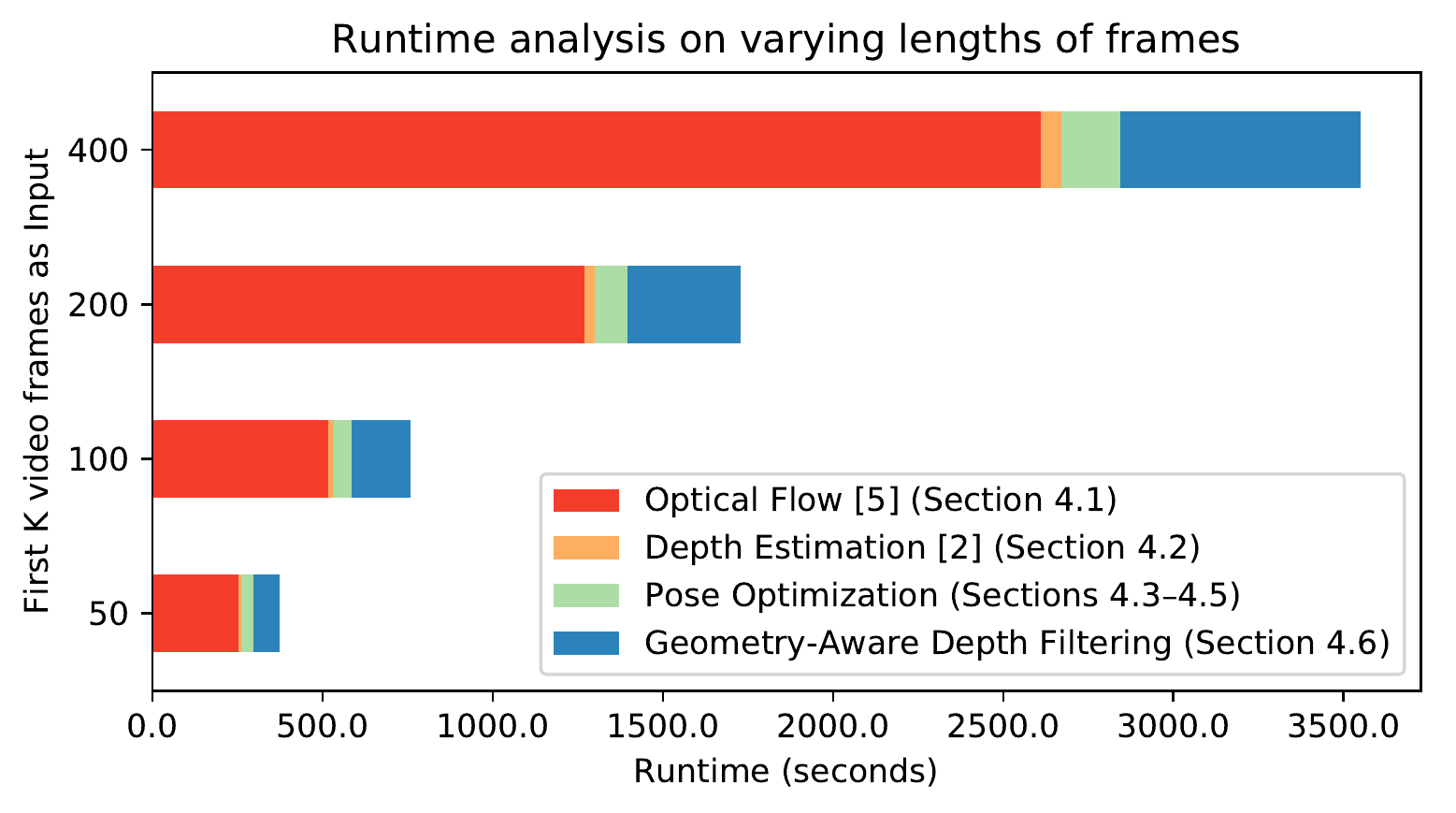}\\%
\caption{\textbf{Runtime analysis of the runtime of different stages in our proposed approach.} }%
\label{fig:ablation}
\end{figure}%

%% file: table/scannet.tex
\begin{table}[t!]
    \setlength{\tabcolsep}{2pt}
    \centering
    \caption{
    \changed{Quantitative comparison on the ScanNet dataset \cite{dai2017scannet} using the test split provided by Tang and Tan~\cite{tang2018ba}.}
    }
    \label{tab:quan_ScanNet}
    \resizebox{8cm}{!}{
      \changed{
    \begin{tabular}{lccccc}
    \toprule
        & \multicolumn{4}{c}{Error metric $\downarrow$} \\
        \cmidrule(lr){2-6}
        & Abs Rel & Sq Rel & RMSE & RMSE log & Sc Inv\\
    \midrule
    DeMoN \cite{ummenhofer2017demon} & 0.231 & 0.520 & 0.761 & 0.289 & 0.284\\
    BA-Net \cite{tang2018ba} & 0.161 & 0.092 & 0.346 & 0.214 & 0.184\\
    DeepV2D (NYU) \cite{teed2020deepv2d} & 0.080 & \ul{0.018} & 0.223 & 0.109 & 0.105\\
    DeepV2D (ScanNet) \cite{teed2020deepv2d} & \textbf{0.057} & \textbf{0.010} & \textbf{0.168} & \textbf{0.080} & \textbf{0.077}\\
    MiDaS-v2 \cite{li2019learning} & 0.208 & 0.318 & 0.742 & 0.246 & 0.239\\
    CVD \cite{Luo-VideoDepth-2020} & \ul{0.073} & 0.037 & \ul{0.217} & \ul{0.105} & \ul{0.103}\\
    Ours & 0.180 & 0.188 & 0.603 & 0.218 & 0.217\\
    \bottomrule
    \end{tabular}
      }
    }
    \label{tab:scannet}
\end{table}

%% file: figures/sintel_depth_partial.tex
\begin{figure}[h!]
\centering
\captionsetup[subfigure]{labelformat=empty}

\begin{subfigure}{.48\linewidth}
\centering
\captionsetup{width=0.98\linewidth}
\includegraphics[width=0.98\linewidth]{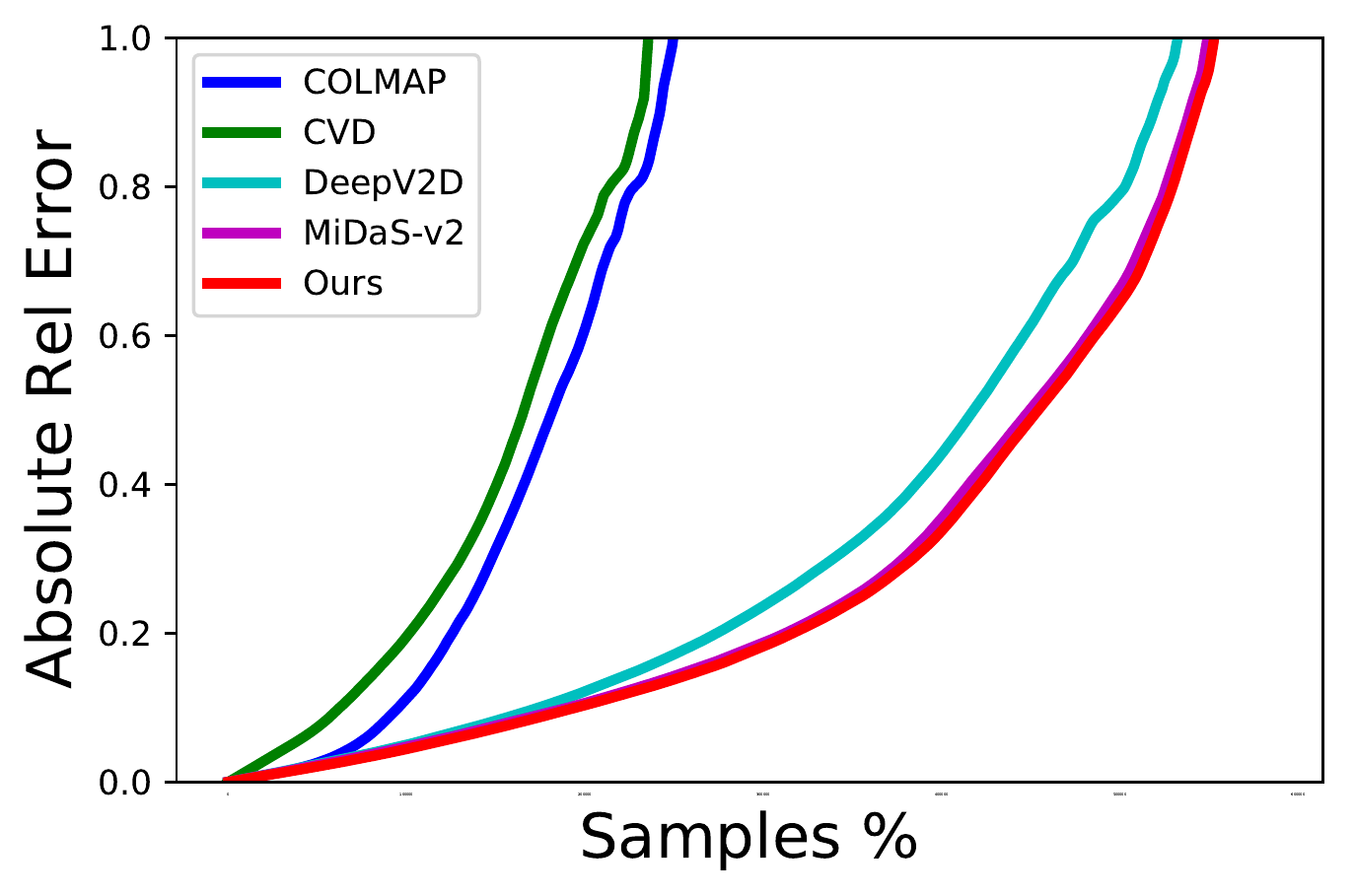}\vspace{-2.5mm}
\caption{{\small (a) Abs Rel - Sintel Clean}}
\end{subfigure}
\begin{subfigure}{.48\linewidth}
\centering
\captionsetup{width=0.98\linewidth}
\includegraphics[width=0.98\linewidth]{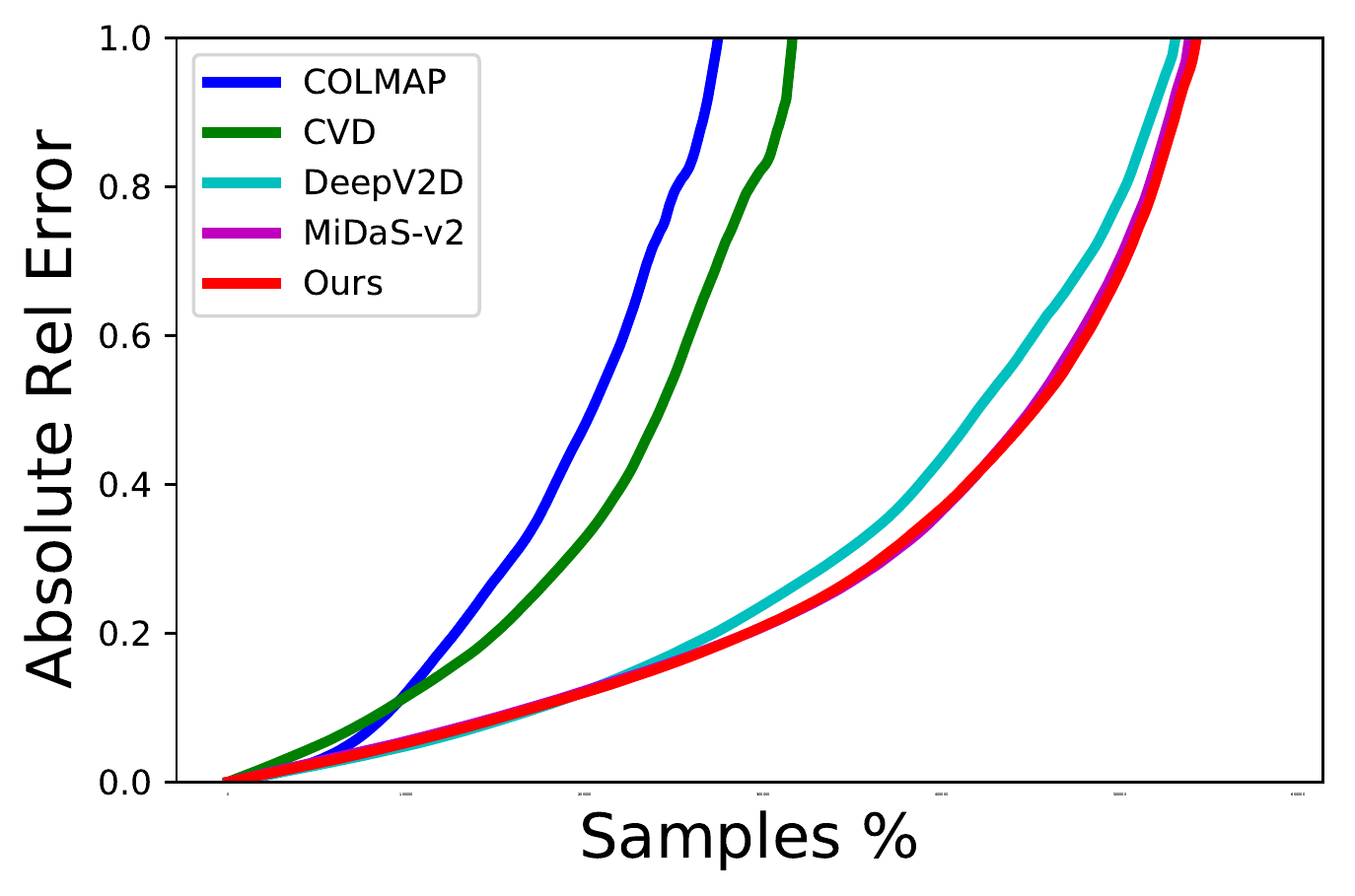}\vspace{-2.5mm}
\caption{{\small (b) Abs Rel - Sintel Final}}
\end{subfigure}

\par\medskip

\caption{\small Comparisons of Absolute Rel Error on Sintel benchmark. (\textit{Clean} and \textit{Final} categories respectively). All the per-frame errors are stored and sorted for plotting the distributions. Note that COLMAP and CVD (which relies on COLMAP) fail on many Sintel sequences, resulting in partial data points.}
\label{fig:sintel_depth_partial}
\end{figure}

%% file: table/sintel_eval_seq.tex
\begin{table*}[t]
\centering
\caption{
\textbf{Per-sequence quantitative evaluations of depth and pose on the MPI Sintel benchmark} (\textit{Sintel Final}). 
}
\vspace{1.5mm}
\label{tab:sintel_depth}
% \mpage{0.98}{
\small \vspace{1em}
\resizebox{\linewidth}{!}{
\renewcommand{\arraystretch}{1} % Default value: 1
\begin{tabular}{lccccccccccccc}
\toprule
& \multicolumn{4}{c}{\texttt{Depth - Error metric} $\downarrow$} && \multicolumn{3}{c}{\texttt{Depth - Accuracy metric} $\uparrow$} &&
\multicolumn{3}{c}{\texttt{Pose - Error metric} $\downarrow$}\\
\cmidrule{2-5} \cmidrule{7-9} \cmidrule{11-13}
Method  &  \textbf{Abs Rel}  & \textbf{Sq Rel}  & \textbf{RMSE}  & \textbf{log RMSE}  && $\mathbf{\delta < 1.25}$ & $\mathbf{\delta < 1.25^2}$ & $\mathbf{\delta < 1.25^3}$ && \textbf{ATE (m)$\downarrow$} & \textbf{RPE Trans (m)$\downarrow$} & \textbf{RPE Rot (deg)$\downarrow$}\\
\midrule
&&&&&& \tb{alley\_1}  &&&&&&\\
DeepV2D~\cite{teed2020deepv2d} &   1.274  &   4.724  &   4.145  &   0.908  &&   0.672  &   0.738  &   0.752  &&   0.103  &   0.043  &   0.023  \\
Ours - Single-scale pose (aligned MiDaS) &   0.511  &   0.969  &   3.146  &   0.512  &&   0.602  &   0.722  &   0.769  &&   0.112  &   0.071  &   0.023  \\
Ours - Single-scale pose + depth fine-tuning &   1.077  &   3.499  &   3.784  &   0.832  &&   0.715  &   0.751  &   0.753  &&   0.033  &   0.012  &   0.006  \\
Ours - Single-scale pose + depth filter &   0.511  &   0.956  &   3.135  &   0.512  &&   0.600  &   0.731  &   0.770  &&   0.112  &   0.071  &   0.023  \\
Ours - Flexible pose &   0.532  &   1.170  &   3.252  &   0.522  &&   0.621  &   0.740  &   0.809  &&   0.026  &   0.015  &   0.002  \\
Ours - Flexible pose + depth fine-tuning &   0.874  &   2.402  &   3.807  &   0.751  &&   0.707  &   0.751  &   0.753  &&   0.026  &   0.010  &   0.003  \\
Ours - Flexible pose + depth filter &   0.531  &   1.157  &   3.250  &   0.521  &&   0.625  &   0.740  &   0.808  &&   0.026  &   0.015  &   0.002  \\
\midrule
&&&&&& \tb{alley\_2}  &&&&&&\\
DeepV2D~\cite{teed2020deepv2d} &   0.317  &   2.324  &   7.561  &   0.560  &&   0.485  &   0.736  &   0.825  &&   0.807  &   0.382  &   0.158  \\
Ours - Single-scale pose (aligned MiDaS) &   0.263  &   1.377  &   5.988  &   0.382  &&   0.556  &   0.806  &   0.918  &&   0.327  &   0.082  &   0.021  \\
Ours - Single-scale pose + depth fine-tuning &   0.379  &   2.423  &   7.646  &   0.594  &&   0.387  &   0.624  &   0.825  &&   0.360  &   0.073  &   0.011  \\
Ours - Single-scale pose + depth filter &   0.261  &   1.370  &   5.996  &   0.381  &&   0.557  &   0.810  &   0.917  &&   0.327  &   0.082  &   0.021  \\
Ours - Flexible pose &   0.244  &   1.332  &   5.987  &   0.370  &&   0.562  &   0.859  &   0.918  &&   0.320  &   0.072  &   0.009  \\
Ours - Flexible pose + depth fine-tuning &   0.292  &   1.820  &   6.878  &   0.465  &&   0.480  &   0.787  &   0.893  &&   0.324  &   0.073  &   0.009  \\
Ours - Flexible pose + depth filter &   0.244  &   1.332  &   5.990  &   0.370  &&   0.563  &   0.860  &   0.918  &&   0.320  &   0.072  &   0.009  \\
\midrule
&&&&&& \tb{ambush\_2}  &&&&&&\\
DeepV2D~\cite{teed2020deepv2d} &   0.716  &   8.314  &  20.149  &   1.764  &&   0.340  &   0.460  &   0.570  &&   1.731  &   3.314  &   0.680  \\
Ours - Single-scale pose (aligned MiDaS) &   0.659  &   7.949  &  19.914  &   1.528  &&   0.365  &   0.497  &   0.611  &&   0.348  &   2.426  &   0.140  \\
Ours - Single-scale pose + depth fine-tuning &   0.702  &   8.303  &  20.140  &   1.747  &&   0.349  &   0.464  &   0.581  &&   0.311  &   2.434  &   0.132  \\
Ours - Single-scale pose + depth filter &   0.656  &   7.947  &  19.921  &   1.532  &&   0.365  &   0.496  &   0.613  &&   0.358  &   2.424  &   0.142  \\
Ours - Flexible pose &   0.657  &   7.958  &  19.923  &   1.532  &&   0.361  &   0.497  &   0.610  &&   0.348  &   2.468  &   0.112  \\
Ours - Flexible pose + depth fine-tuning &   0.686  &   8.125  &  20.049  &   1.648  &&   0.351  &   0.467  &   0.582  &&   0.385  &   2.466  &   0.123  \\
Ours - Flexible pose + depth filter &   0.656  &   7.953  &  19.927  &   1.533  &&   0.360  &   0.496  &   0.611  &&   0.348  &   2.468  &   0.112  \\
\midrule
&&&&&& \tb{ambush\_4}  &&&&&&\\
DeepV2D~\cite{teed2020deepv2d} &   0.781  &   2.307  &   2.510  &   0.758  &&   0.357  &   0.479  &   0.649  &&   1.629  &   1.638  &   1.011  \\
Ours - Single-scale pose (aligned MiDaS) &   0.610  &   1.566  &   2.483  &   0.651  &&   0.335  &   0.510  &   0.697  &&   0.240  &   0.285  &   0.071  \\
Ours - Single-scale pose + depth fine-tuning &   0.688  &   1.883  &   2.436  &   0.713  &&   0.371  &   0.478  &   0.685  &&   0.124  &   0.226  &   0.030  \\
Ours - Single-scale pose + depth filter &   0.593  &   1.463  &   2.447  &   0.646  &&   0.342  &   0.515  &   0.709  &&   0.242  &   0.285  &   0.079  \\
Ours - Flexible pose &   0.605  &   1.462  &   2.446  &   0.655  &&   0.320  &   0.520  &   0.706  &&   0.199  &   0.244  &   0.050  \\
Ours - Flexible pose + depth fine-tuning &   0.625  &   1.585  &   2.434  &   0.688  &&   0.373  &   0.512  &   0.679  &&   0.188  &   0.236  &   0.038  \\
Ours - Flexible pose + depth filter &   0.601  &   1.442  &   2.436  &   0.654  &&   0.323  &   0.523  &   0.708  &&   0.199  &   0.244  &   0.050  \\
\midrule
&&&&&& \tb{ambush\_5}  &&&&&&\\
DeepV2D~\cite{teed2020deepv2d} &   0.416  &   7.732  &  19.769  &   1.286  &&   0.363  &   0.642  &   0.732  &&   1.294  &   1.240  &   0.962  \\
Ours - Single-scale pose (aligned MiDaS) &   0.335  &   7.329  &  19.421  &   1.158  &&   0.548  &   0.727  &   0.756  &&   0.277  &   0.333  &   0.038  \\
Ours - Single-scale pose + depth fine-tuning &   0.354  &   7.807  &  19.907  &   1.332  &&   0.517  &   0.718  &   0.745  &&   0.165  &   0.304  &   0.031  \\
Ours - Single-scale pose + depth filter &   0.330  &   7.331  &  19.428  &   1.157  &&   0.561  &   0.732  &   0.757  &&   0.291  &   0.329  &   0.046  \\
Ours - Flexible pose &   0.345  &   7.388  &  19.483  &   1.178  &&   0.518  &   0.708  &   0.753  &&   0.199  &   0.298  &   0.030  \\
Ours - Flexible pose + depth fine-tuning &   0.397  &   7.869  &  19.903  &   1.346  &&   0.413  &   0.673  &   0.746  &&   0.200  &   0.312  &   0.027  \\
Ours - Flexible pose + depth filter &   0.339  &   7.390  &  19.490  &   1.176  &&   0.530  &   0.715  &   0.754  &&   0.199  &   0.298  &   0.030  \\
\midrule
&&&&&& \tb{ambush\_6}  &&&&&&\\
DeepV2D~\cite{teed2020deepv2d} &   0.615  &   1.071  &   2.316  &   0.681  &&   0.385  &   0.595  &   0.703  &&   2.603  &   2.576  &   1.058  \\
Ours - Single-scale pose (aligned MiDaS) &   0.315  &   0.725  &   2.342  &   0.458  &&   0.543  &   0.749  &   0.880  &&   0.292  &   1.266  &   0.071  \\
Ours - Single-scale pose + depth fine-tuning &   0.535  &   0.919  &   2.278  &   0.643  &&   0.399  &   0.609  &   0.722  &&   0.230  &   1.287  &   0.068  \\
Ours - Single-scale pose + depth filter &   0.302  &   0.689  &   2.320  &   0.449  &&   0.566  &   0.754  &   0.888  &&   0.236  &   1.257  &   0.079  \\
Ours - Flexible pose &   0.323  &   0.742  &   2.367  &   0.471  &&   0.541  &   0.751  &   0.879  &&   0.197  &   1.274  &   0.062  \\
Ours - Flexible pose + depth fine-tuning &   0.433  &   0.710  &   2.191  &   0.563  &&   0.433  &   0.650  &   0.784  &&   0.287  &   1.279  &   0.072  \\
Ours - Flexible pose + depth filter &   0.317  &   0.724  &   2.358  &   0.466  &&   0.551  &   0.755  &   0.880  &&   0.197  &   1.274  &   0.062  \\
\midrule
&&&&&& \tb{ambush\_7}  &&&&&&\\
DeepV2D~\cite{teed2020deepv2d} &   0.064  &   0.005  &   0.053  &   0.089  &&   0.973  &   0.999  &   1.000  &&   0.632  &   0.290  &   0.193  \\
Ours - Single-scale pose (aligned MiDaS) &   0.125  &   0.014  &   0.085  &   0.148  &&   0.847  &   0.995  &   1.000  &&   0.208  &   0.053  &   0.035  \\
Ours - Single-scale pose + depth fine-tuning &   0.051  &   0.003  &   0.040  &   0.067  &&   0.997  &   1.000  &   1.000  &&   0.120  &   0.019  &   0.014  \\
Ours - Single-scale pose + depth filter &   0.120  &   0.012  &   0.080  &   0.142  &&   0.865  &   0.999  &   1.000  &&   0.160  &   0.049  &   0.031  \\
Ours - Flexible pose &   0.108  &   0.009  &   0.071  &   0.136  &&   0.896  &   0.999  &   1.000  &&   0.091  &   0.022  &   0.011  \\
Ours - Flexible pose + depth fine-tuning &   0.081  &   0.005  &   0.056  &   0.104  &&   0.983  &   1.000  &   1.000  &&   0.202  &   0.031  &   0.015  \\
Ours - Flexible pose + depth filter &   0.106  &   0.009  &   0.069  &   0.132  &&   0.912  &   0.999  &   1.000  &&   0.091  &   0.022  &   0.011  \\
\midrule
&&&&&& \tb{bamboo\_1}  &&&&&&\\
DeepV2D~\cite{teed2020deepv2d} &   0.513  &   2.243  &   2.192  &   0.519  &&   0.696  &   0.788  &   0.842  &&   0.729  &   0.259  &   0.141  \\
Ours - Single-scale pose (aligned MiDaS) &   0.539  &   2.368  &   2.351  &   0.532  &&   0.686  &   0.781  &   0.834  &&   0.133  &   0.075  &   0.026  \\
Ours - Single-scale pose + depth fine-tuning &   0.493  &   2.051  &   2.124  &   0.507  &&   0.684  &   0.793  &   0.844  &&   0.079  &   0.019  &   0.013  \\
Ours - Single-scale pose + depth filter &   0.536  &   2.362  &   2.339  &   0.531  &&   0.692  &   0.781  &   0.834  &&   0.133  &   0.075  &   0.026  \\
Ours - Flexible pose &   0.477  &   1.955  &   2.085  &   0.501  &&   0.707  &   0.791  &   0.845  &&   0.121  &   0.018  &   0.004  \\
Ours - Flexible pose + depth fine-tuning &   0.494  &   2.070  &   2.123  &   0.509  &&   0.698  &   0.790  &   0.844  &&   0.097  &   0.016  &   0.006  \\
Ours - Flexible pose + depth filter &   0.478  &   1.960  &   2.088  &   0.501  &&   0.707  &   0.790  &   0.845  &&   0.121  &   0.018  &   0.004  \\
\midrule
&&&&&& \tb{bamboo\_2}  &&&&&&\\
DeepV2D~\cite{teed2020deepv2d} &   0.518  &   1.792  &   4.190  &   0.636  &&   0.350  &   0.602  &   0.766  &&   0.431  &   0.125  &   0.083  \\
Ours - Single-scale pose (aligned MiDaS) &   0.428  &   1.433  &   4.041  &   0.572  &&   0.400  &   0.653  &   0.806  &&   0.110  &   0.067  &   0.024  \\
Ours - Single-scale pose + depth fine-tuning &   0.482  &   1.622  &   4.255  &   0.627  &&   0.294  &   0.612  &   0.778  &&   0.018  &   0.020  &   0.011  \\
Ours - Single-scale pose + depth filter &   0.426  &   1.426  &   4.040  &   0.570  &&   0.408  &   0.654  &   0.809  &&   0.110  &   0.067  &   0.024  \\
Ours - Flexible pose &   0.415  &   1.348  &   3.975  &   0.552  &&   0.386  &   0.688  &   0.821  &&   0.040  &   0.014  &   0.003  \\
Ours - Flexible pose + depth fine-tuning &   0.494  &   1.656  &   4.240  &   0.629  &&   0.303  &   0.596  &   0.775  &&   0.015  &   0.016  &   0.006  \\
Ours - Flexible pose + depth filter &   0.415  &   1.348  &   3.977  &   0.552  &&   0.387  &   0.688  &   0.821  &&   0.040  &   0.014  &   0.003  \\
\bottomrule
\end{tabular}
}
% \vspace{-4mm}
\label{tab:sintel_eval_seq_1}
\end{table*}

\begin{table*}[t]
\centering
\caption{
\textbf{Per-sequence quantitative evaluations of depth and pose on the MPI Sintel benchmark} (\textit{Sintel Final}). 
}
\vspace{1.5mm}
\label{tab:sintel_depth}
% \mpage{0.98}{
\small \vspace{1em}
\resizebox{\linewidth}{!}{
\renewcommand{\arraystretch}{1} % Default value: 1
\begin{tabular}{lccccccccccccc}
\toprule
& \multicolumn{4}{c}{\texttt{Depth - Error metric} $\downarrow$} && \multicolumn{3}{c}{\texttt{Depth - Accuracy metric} $\uparrow$} &&
\multicolumn{3}{c}{\texttt{Pose - Error metric} $\downarrow$}\\
\cmidrule{2-5} \cmidrule{7-9} \cmidrule{11-13}
Method  &  \textbf{Abs Rel}  & \textbf{Sq Rel}  & \textbf{RMSE}  & \textbf{log RMSE}  && $\mathbf{\delta < 1.25}$ & $\mathbf{\delta < 1.25^2}$ & $\mathbf{\delta < 1.25^3}$ && \textbf{ATE (m)$\downarrow$} & \textbf{RPE Trans (m)$\downarrow$} & \textbf{RPE Rot (deg)$\downarrow$}\\
\midrule

&&&&&& \tb{bandage\_1}  &&&&&&\\
DeepV2D~\cite{teed2020deepv2d} &   0.214  &   0.084  &   0.334  &   0.329  &&   0.591  &   0.845  &   0.922  &&   0.762  &   0.425  &   0.255  \\
Ours - Single-scale pose (aligned MiDaS) &   0.864  &   3.635  &   1.714  &   0.674  &&   0.624  &   0.712  &   0.776  &&   0.077  &   0.034  &   0.002  \\
Ours - Single-scale pose + depth fine-tuning &   0.146  &   0.051  &   0.272  &   0.241  &&   0.765  &   0.883  &   0.990  &&   0.083  &   0.016  &   0.007  \\
Ours - Single-scale pose + depth filter &   0.857  &   3.418  &   1.694  &   0.675  &&   0.625  &   0.709  &   0.775  &&   0.077  &   0.034  &   0.002  \\
Ours - Flexible pose &   0.850  &   3.352  &   1.676  &   0.673  &&   0.624  &   0.714  &   0.779  &&   0.070  &   0.026  &   0.002  \\
Ours - Flexible pose + depth fine-tuning &   0.125  &   0.040  &   0.240  &   0.208  &&   0.825  &   0.912  &   0.997  &&   0.077  &   0.011  &   0.005  \\
Ours - Flexible pose + depth filter &   0.845  &   3.210  &   1.661  &   0.674  &&   0.625  &   0.711  &   0.780  &&   0.070  &   0.026  &   0.002  \\
\midrule
&&&&&& \tb{bandage\_2}  &&&&&&\\
DeepV2D~\cite{teed2020deepv2d} &   0.438  &   0.231  &   0.474  &   0.469  &&   0.260  &   0.627  &   0.872  &&   0.062  &   0.019  &   0.007  \\
Ours - Single-scale pose (aligned MiDaS) &   0.550  &   1.135  &   1.255  &   0.527  &&   0.400  &   0.631  &   0.796  &&   0.103  &   0.042  &   0.010  \\
Ours - Single-scale pose + depth fine-tuning &   0.418  &   0.184  &   0.412  &   0.424  &&   0.370  &   0.612  &   0.909  &&   0.034  &   0.018  &   0.006  \\
Ours - Single-scale pose + depth filter &   0.542  &   1.066  &   1.229  &   0.522  &&   0.399  &   0.631  &   0.799  &&   0.106  &   0.045  &   0.010  \\
Ours - Flexible pose &   0.492  &   0.668  &   0.960  &   0.513  &&   0.346  &   0.591  &   0.801  &&   0.082  &   0.022  &   0.003  \\
Ours - Flexible pose + depth fine-tuning &   0.470  &   0.241  &   0.487  &   0.484  &&   0.229  &   0.559  &   0.864  &&   0.045  &   0.020  &   0.004  \\
Ours - Flexible pose + depth filter &   0.488  &   0.644  &   0.940  &   0.511  &&   0.347  &   0.584  &   0.805  &&   0.082  &   0.022  &   0.003  \\
\midrule
&&&&&& \tb{cave\_2}  &&&&&&\\
DeepV2D~\cite{teed2020deepv2d} &   0.972  &  16.038  &  18.036  &   0.930  &&   0.281  &   0.435  &   0.545  &&   1.589  &   3.550  &   1.114  \\
Ours - Single-scale pose (aligned MiDaS) &   0.700  &  10.207  &  15.807  &   0.730  &&   0.312  &   0.524  &   0.651  &&   0.941  &   3.034  &   0.081  \\
Ours - Single-scale pose + depth fine-tuning &   0.904  &  14.897  &  17.791  &   0.894  &&   0.301  &   0.448  &   0.569  &&   0.879  &   3.036  &   0.087  \\
Ours - Single-scale pose + depth filter &   0.688  &  10.371  &  15.901  &   0.725  &&   0.317  &   0.530  &   0.654  &&   1.113  &   3.041  &   0.081  \\
Ours - Flexible pose &   0.689  &   9.851  &  15.988  &   0.735  &&   0.296  &   0.522  &   0.658  &&   0.968  &   3.027  &   0.083  \\
Ours - Flexible pose + depth fine-tuning &   0.810  &  11.986  &  17.659  &   0.877  &&   0.324  &   0.473  &   0.573  &&   0.789  &   3.020  &   0.082  \\
Ours - Flexible pose + depth filter &   0.683  &   9.686  &  15.969  &   0.732  &&   0.299  &   0.528  &   0.659  &&   0.968  &   3.027  &   0.083  \\
\midrule
&&&&&& \tb{cave\_4}  &&&&&&\\
DeepV2D~\cite{teed2020deepv2d} &   0.444  &   2.244  &   5.513  &   0.631  &&   0.355  &   0.595  &   0.733  &&   0.936  &   1.530  &   0.840  \\
Ours - Single-scale pose (aligned MiDaS) &   0.365  &   1.662  &   4.868  &   0.524  &&   0.411  &   0.660  &   0.805  &&   0.188  &   0.977  &   0.044  \\
Ours - Single-scale pose + depth fine-tuning &   0.406  &   2.041  &   5.403  &   0.602  &&   0.375  &   0.631  &   0.759  &&   0.177  &   0.962  &   0.030  \\
Ours - Single-scale pose + depth filter &   0.359  &   1.632  &   4.855  &   0.518  &&   0.413  &   0.666  &   0.807  &&   0.184  &   0.977  &   0.045  \\
Ours - Flexible pose &   0.356  &   1.684  &   4.965  &   0.529  &&   0.426  &   0.673  &   0.797  &&   0.205  &   0.960  &   0.030  \\
Ours - Flexible pose + depth fine-tuning &   0.425  &   2.135  &   5.449  &   0.613  &&   0.372  &   0.615  &   0.748  &&   0.167  &   0.948  &   0.047  \\
Ours - Flexible pose + depth filter &   0.353  &   1.669  &   4.951  &   0.526  &&   0.429  &   0.676  &   0.797  &&   0.205  &   0.960  &   0.030  \\
\midrule
&&&&&& \tb{market\_2}  &&&&&&\\
DeepV2D~\cite{teed2020deepv2d} &   0.613  &  10.914  &  18.792  &   0.648  &&   0.307  &   0.499  &   0.662  &&   0.433  &   0.233  &   0.169  \\
Ours - Single-scale pose (aligned MiDaS) &   0.254  &   3.281  &  12.238  &   0.336  &&   0.564  &   0.815  &   0.954  &&   0.089  &   0.051  &   0.017  \\
Ours - Single-scale pose + depth fine-tuning &   0.459  &   7.052  &  16.473  &   0.530  &&   0.388  &   0.612  &   0.735  &&   0.042  &   0.018  &   0.008  \\
Ours - Single-scale pose + depth filter &   0.252  &   3.235  &  12.204  &   0.333  &&   0.565  &   0.817  &   0.957  &&   0.089  &   0.051  &   0.017  \\
Ours - Flexible pose &   0.345  &   4.641  &  13.972  &   0.413  &&   0.388  &   0.720  &   0.887  &&   0.045  &   0.020  &   0.003  \\
Ours - Flexible pose + depth fine-tuning &   0.507  &   8.346  &  18.137  &   0.589  &&   0.309  &   0.530  &   0.699  &&   0.054  &   0.016  &   0.006  \\
Ours - Flexible pose + depth filter &   0.344  &   4.598  &  13.963  &   0.412  &&   0.384  &   0.715  &   0.889  &&   0.045  &   0.020  &   0.003  \\
\midrule
&&&&&& \tb{market\_5}  &&&&&&\\
DeepV2D~\cite{teed2020deepv2d} &   0.780  &   2.505  &   4.181  &   0.846  &&   0.246  &   0.435  &   0.560  &&   2.464  &   1.406  &   1.048  \\
Ours - Single-scale pose (aligned MiDaS) &   0.418  &   1.253  &   3.428  &   0.565  &&   0.385  &   0.634  &   0.782  &&   1.974  &   0.316  &   0.053  \\
Ours - Single-scale pose + depth fine-tuning &   0.666  &   2.062  &   4.053  &   0.776  &&   0.268  &   0.467  &   0.604  &&   1.982  &   0.277  &   0.036  \\
Ours - Single-scale pose + depth filter &   0.432  &   1.247  &   3.409  &   0.559  &&   0.378  &   0.616  &   0.775  &&   1.768  &   0.326  &   0.056  \\
Ours - Flexible pose &   0.417  &   1.230  &   3.375  &   0.560  &&   0.408  &   0.625  &   0.781  &&   1.986  &   0.308  &   0.032  \\
Ours - Flexible pose + depth fine-tuning &   0.517  &   1.784  &   4.026  &   0.734  &&   0.306  &   0.551  &   0.690  &&   1.943  &   0.281  &   0.032  \\
Ours - Flexible pose + depth filter &   0.432  &   1.224  &   3.352  &   0.552  &&   0.392  &   0.608  &   0.776  &&   1.986  &   0.308  &   0.032  \\
\midrule
&&&&&& \tb{market\_6}  &&&&&&\\
DeepV2D~\cite{teed2020deepv2d} &   0.951  &   6.073  &  12.243  &   1.131  &&   0.184  &   0.293  &   0.409  &&   1.425  &   1.374  &   0.993  \\
Ours - Single-scale pose (aligned MiDaS) &   0.655  &   4.255  &  10.912  &   0.850  &&   0.234  &   0.377  &   0.541  &&   0.715  &   0.359  &   0.033  \\
Ours - Single-scale pose + depth fine-tuning &   0.757  &   5.154  &  11.906  &   1.003  &&   0.206  &   0.324  &   0.467  &&   0.717  &   0.344  &   0.024  \\
Ours - Single-scale pose + depth filter &   0.649  &   4.230  &  10.916  &   0.846  &&   0.236  &   0.378  &   0.542  &&   0.715  &   0.359  &   0.033  \\
Ours - Flexible pose &   0.590  &   3.976  &  10.708  &   0.804  &&   0.241  &   0.400  &   0.634  &&   0.695  &   0.349  &   0.016  \\
Ours - Flexible pose + depth fine-tuning &   0.620  &   4.822  &  11.863  &   0.952  &&   0.227  &   0.378  &   0.570  &&   0.612  &   0.352  &   0.026  \\
Ours - Flexible pose + depth filter &   0.587  &   3.965  &  10.711  &   0.803  &&   0.242  &   0.399  &   0.639  &&   0.695  &   0.349  &   0.016  \\
\midrule
&&&&&& \tb{shaman\_2}  &&&&&&\\
DeepV2D~\cite{teed2020deepv2d} &   0.326  &   0.639  &   1.504  &   0.715  &&   0.549  &   0.568  &   0.570  &&   0.016  &   0.002  &   0.002  \\
Ours - Single-scale pose (aligned MiDaS) &   0.152  &   0.188  &   0.813  &   0.362  &&   0.723  &   0.858  &   0.946  &&   0.074  &   0.025  &   0.006  \\
Ours - Single-scale pose + depth fine-tuning &   0.349  &   0.703  &   1.568  &   0.783  &&   0.561  &   0.568  &   0.569  &&   0.068  &   0.014  &   0.005  \\
Ours - Single-scale pose + depth filter &   0.151  &   0.186  &   0.809  &   0.360  &&   0.724  &   0.863  &   0.948  &&   0.074  &   0.025  &   0.006  \\
Ours - Flexible pose &   0.153  &   0.187  &   0.812  &   0.361  &&   0.723  &   0.858  &   0.947  &&   0.062  &   0.014  &   0.003  \\
Ours - Flexible pose + depth fine-tuning &   0.189  &   0.247  &   0.957  &   0.386  &&   0.612  &   0.833  &   0.913  &&   0.047  &   0.006  &   0.002  \\
Ours - Flexible pose + depth filter &   0.152  &   0.186  &   0.809  &   0.359  &&   0.724  &   0.862  &   0.948  &&   0.062  &   0.014  &   0.003  \\
\midrule
&&&&&& \tb{shaman\_3}  &&&&&&\\
DeepV2D~\cite{teed2020deepv2d} &   0.210  &   0.028  &   0.113  &   0.229  &&   0.614  &   0.970  &   1.000  &&   0.417  &   0.185  &   0.102  \\
Ours - Single-scale pose (aligned MiDaS) &   0.194  &   0.038  &   0.137  &   0.220  &&   0.681  &   0.949  &   0.999  &&   0.086  &   0.055  &   0.014  \\
Ours - Single-scale pose + depth fine-tuning &   0.142  &   0.013  &   0.082  &   0.161  &&   0.846  &   1.000  &   1.000  &&   0.040  &   0.011  &   0.004  \\
Ours - Single-scale pose + depth filter &   0.190  &   0.036  &   0.135  &   0.216  &&   0.689  &   0.957  &   0.999  &&   0.086  &   0.055  &   0.014  \\
Ours - Flexible pose &   0.199  &   0.039  &   0.143  &   0.222  &&   0.674  &   0.963  &   0.996  &&   0.054  &   0.019  &   0.003  \\
Ours - Flexible pose + depth fine-tuning &   0.125  &   0.011  &   0.072  &   0.147  &&   0.853  &   1.000  &   1.000  &&   0.040  &   0.010  &   0.003  \\
Ours - Flexible pose + depth filter &   0.197  &   0.038  &   0.142  &   0.221  &&   0.680  &   0.959  &   0.999  &&   0.054  &   0.019  &   0.003  \\
\bottomrule
\end{tabular}
}
% \vspace{-4mm}
\label{tab:sintel_eval_seq_2}
\end{table*}

\begin{table*}[t]
\centering
\caption{
\textbf{Per-sequence quantitative evaluations of depth and pose on the MPI Sintel benchmark} (\textit{Sintel Final}). 
}
\vspace{1.5mm}
\label{tab:sintel_depth}
% \mpage{0.98}{
\small \vspace{1em}
\resizebox{\linewidth}{!}{
\renewcommand{\arraystretch}{1} % Default value: 1
\begin{tabular}{lccccccccccccc}
\toprule
& \multicolumn{4}{c}{\texttt{Depth - Error metric} $\downarrow$} && \multicolumn{3}{c}{\texttt{Depth - Accuracy metric} $\uparrow$} &&
\multicolumn{3}{c}{\texttt{Pose - Error metric} $\downarrow$}\\
\cmidrule{2-5} \cmidrule{7-9} \cmidrule{11-13}
Method  &  \textbf{Abs Rel}  & \textbf{Sq Rel}  & \textbf{RMSE}  & \textbf{log RMSE}  && $\mathbf{\delta < 1.25}$ & $\mathbf{\delta < 1.25^2}$ & $\mathbf{\delta < 1.25^3}$ && \textbf{ATE (m)$\downarrow$} & \textbf{RPE Trans (m)$\downarrow$} & \textbf{RPE Rot (deg)$\downarrow$}\\
\midrule

&&&&&& \tb{sleeping\_1}  &&&&&&\\
DeepV2D~\cite{teed2020deepv2d} &   0.060  &   0.005  &   0.059  &   0.081  &&   0.980  &   1.000  &   1.000  &&   0.011  &   0.009  &   0.008  \\
Ours - Single-scale pose (aligned MiDaS) &   0.129  &   0.027  &   0.141  &   0.154  &&   0.907  &   0.983  &   0.996  &&   0.091  &   0.033  &   0.019  \\
Ours - Single-scale pose + depth fine-tuning &   0.059  &   0.004  &   0.053  &   0.073  &&   0.999  &   1.000  &   1.000  &&   0.080  &   0.011  &   0.007  \\
Ours - Single-scale pose + depth filter &   0.128  &   0.026  &   0.139  &   0.152  &&   0.914  &   0.983  &   0.996  &&   0.091  &   0.033  &   0.019  \\
Ours - Flexible pose &   0.135  &   0.033  &   0.155  &   0.161  &&   0.911  &   0.980  &   0.990  &&   0.090  &   0.011  &   0.006  \\
Ours - Flexible pose + depth fine-tuning &   0.056  &   0.004  &   0.053  &   0.074  &&   0.995  &   1.000  &   1.000  &&   0.075  &   0.008  &   0.003  \\
Ours - Flexible pose + depth filter &   0.134  &   0.033  &   0.154  &   0.161  &&   0.913  &   0.980  &   0.990  &&   0.090  &   0.011  &   0.006  \\
\midrule
&&&&&& \tb{sleeping\_2}  &&&&&&\\
DeepV2D~\cite{teed2020deepv2d} &   0.219  &   0.119  &   0.495  &   0.238  &&   0.525  &   0.985  &   1.000  &&   0.134  &   0.010  &   0.008  \\
Ours - Single-scale pose (aligned MiDaS) &   0.279  &   0.193  &   0.602  &   0.288  &&   0.400  &   0.913  &   1.000  &&   0.161  &   0.049  &   0.024  \\
Ours - Single-scale pose + depth fine-tuning &   0.280  &   0.193  &   0.590  &   0.296  &&   0.462  &   0.870  &   1.000  &&   0.169  &   0.016  &   0.004  \\
Ours - Single-scale pose + depth filter &   0.278  &   0.191  &   0.599  &   0.287  &&   0.396  &   0.913  &   1.000  &&   0.161  &   0.049  &   0.024  \\
Ours - Flexible pose &   0.280  &   0.193  &   0.604  &   0.289  &&   0.391  &   0.923  &   0.999  &&   0.179  &   0.017  &   0.004  \\
Ours - Flexible pose + depth fine-tuning &   0.303  &   0.219  &   0.642  &   0.319  &&   0.358  &   0.839  &   1.000  &&   0.169  &   0.015  &   0.003  \\
Ours - Flexible pose + depth filter &   0.280  &   0.193  &   0.603  &   0.289  &&   0.388  &   0.924  &   0.999  &&   0.179  &   0.017  &   0.004  \\
\midrule
&&&&&& \tb{temple\_2}  &&&&&&\\
DeepV2D~\cite{teed2020deepv2d} &   0.653  &   5.760  &  10.404  &   0.603  &&   0.558  &   0.838  &   0.879  &&   2.638  &   0.965  &   0.740  \\
Ours - Single-scale pose (aligned MiDaS) &   0.614  &   5.168  &  10.410  &   0.582  &&   0.502  &   0.755  &   0.884  &&   2.227  &   0.268  &   0.017  \\
Ours - Single-scale pose + depth fine-tuning &   0.609  &   5.681  &  10.122  &   0.569  &&   0.726  &   0.871  &   0.890  &&   2.245  &   0.259  &   0.022  \\
Ours - Single-scale pose + depth filter &   0.618  &   5.175  &  10.385  &   0.581  &&   0.502  &   0.757  &   0.888  &&   2.227  &   0.268  &   0.017  \\
Ours - Flexible pose &   0.632  &   5.870  &  10.173  &   0.578  &&   0.597  &   0.831  &   0.885  &&   2.323  &   0.265  &   0.015  \\
Ours - Flexible pose + depth fine-tuning &   0.638  &   6.483  &   9.990  &   0.563  &&   0.755  &   0.881  &   0.893  &&   2.286  &   0.259  &   0.019  \\
Ours - Flexible pose + depth filter &   0.632  &   5.827  &  10.152  &   0.577  &&   0.599  &   0.835  &   0.885  &&   2.323  &   0.265  &   0.015  \\
\midrule
&&&&&& \tb{temple\_3}  &&&&&&\\
DeepV2D~\cite{teed2020deepv2d} &   0.481  &   4.485  &   7.312  &   0.686  &&   0.631  &   0.689  &   0.732  &&   2.275  &   2.139  &   1.603  \\
Ours - Single-scale pose (aligned MiDaS) &   0.384  &   3.315  &   6.790  &   0.544  &&   0.606  &   0.729  &   0.807  &&   1.111  &   0.532  &   0.099  \\
Ours - Single-scale pose + depth fine-tuning &   0.450  &   4.174  &   7.225  &   0.650  &&   0.623  &   0.703  &   0.770  &&   0.723  &   0.332  &   0.090  \\
Ours - Single-scale pose + depth filter &   0.377  &   3.188  &   6.720  &   0.541  &&   0.611  &   0.729  &   0.811  &&   1.111  &   0.532  &   0.099  \\
Ours - Flexible pose &   0.410  &   3.430  &   6.807  &   0.554  &&   0.570  &   0.715  &   0.810  &&   0.888  &   0.412  &   0.056  \\
Ours - Flexible pose + depth fine-tuning &   0.466  &   4.613  &   7.341  &   0.661  &&   0.635  &   0.703  &   0.762  &&   0.565  &   0.345  &   0.099  \\
Ours - Flexible pose + depth filter &   0.399  &   3.227  &   6.714  &   0.548  &&   0.584  &   0.719  &   0.816  &&   0.888  &   0.412  &   0.056  \\

\bottomrule
\end{tabular}
}
% \vspace{-4mm}
\label{tab:sintel_eval_seq_3}
\end{table*}